%% file: main.tex
\documentclass{article}
\usepackage[final]{corl_2026} %

\input{preamble}

\title{Intercepting the Future: Latent-Space Predictive World Model for Dynamic VLA Manipulation}

\author{
  Shahram Najam Syed\textsuperscript{1} \\
  \texttt{snsyed@andrew.cmu.edu} \\
  \And
  Arthur Jakobsson\textsuperscript{1}\thanks{Equal contribution.} \\
  \texttt{ajakobss@andrew.cmu.edu} \\
  \And
  Haoran Hao\textsuperscript{1}\footnotemark[\value{footnote}] \\
  \texttt{haoranha@andrew.cmu.edu} \\
  \AND
  Jeffrey Ichnowski\textsuperscript{1} \\
  \texttt{jichnows@andrew.cmu.edu} \\
  \\
  \textsuperscript{1}Robotics Institute, Carnegie Mellon University, Pittsburgh, USA \\
}

\begin{document}

\maketitle

\input{figures/hero}
\begin{abstract}
\input{sections/abstract}
\end{abstract}

\keywords{Vision-Language-Action Models, World Models, Manipulation}

\input{sections/section-1-introduction}
\input{sections/section-2-related-work}
\input{sections/section-3-problem}
\input{sections/section-4-method}
\input{sections/section-5-experiments}
\input{sections/section-6-conclusion}

\newpage

\nocite{*}
\bibliography{references}

\newpage
\appendix
\input{appendix/appendix-A0-Architecture-Pipeline}
\newpage
\input{appendix/appendix-A-method-details}
\newpage
\input{appendix/appendix-B-experimental-setup}
\newpage
\input{appendix/appendix-C-additional-results}
\newpage
\input{appendix/appendix-D-ablation-details}

\end{document}

%% file: preamble.tex
\usepackage{amssymb}
\usepackage{amsmath}
\usepackage{graphicx}
\usepackage[dvipsnames]{xcolor}
\usepackage{tikz}
\usetikzlibrary{arrows.meta,bending,positioning,calc,math,backgrounds,fit}
\usepackage{booktabs}
\usepackage{algorithm}
\usepackage{algorithmicx}
\usepackage{algpseudocode}
\usepackage{enumitem}
\usepackage{multirow}
\usepackage[separate-uncertainty=true]{siunitx}
\usepackage[font=footnotesize]{caption}
\usepackage[font=footnotesize]{subcaption}
\usepackage{tabularx}
\usepackage{xcolor}
\usepackage{epigraph}
\usepackage[most]{tcolorbox}

\newtcolorbox{leftbarquote}{
  enhanced, breakable,
  colback=white, colframe=black,
  boxrule=0pt, frame hidden,
  borderline west={1.5pt}{0pt}{gray!80},
  left=20pt, right=0pt, top=2pt, bottom=2pt,
  sharp corners
}

\newcommand{\methodName}{Anticipatory Horizon Extrapolation with Adaptive Dynamics}
\newcommand{\methodAbbr}{AHEAD}
\usepackage{xcolor}

\newcommand{\inlinecomment}[3]{\textcolor{#1}{[#2 -#3]}}
\renewcommand{\inlinecomment}[3]{}

\usepackage{tikz}

\newcommand{\roundpic}[2][0.245\linewidth]{%
    \tikz[baseline=(img.base)]{
        \node[inner sep=0pt, outer sep=0pt, rounded corners=4pt, clip] (img)
        {\includegraphics[width=#1]{#2}};
    }%
}

%% file: figures/hero.tex
\begin{figure}[htbp]
    \centering
    \setlength{\tabcolsep}{1pt}
    \begin{tabular}{cccc}
        \roundpic{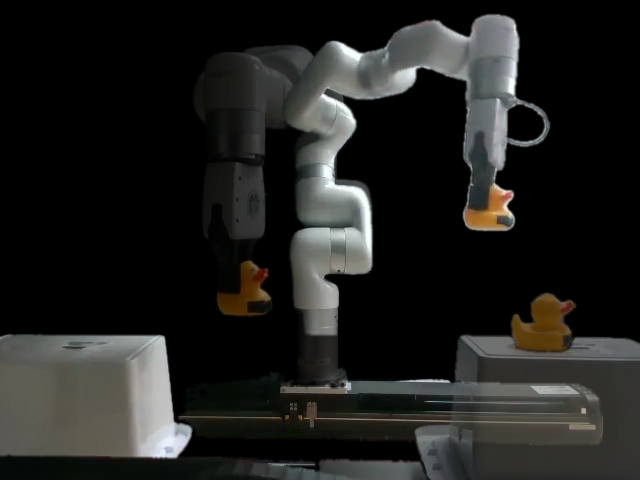} &
        \roundpic{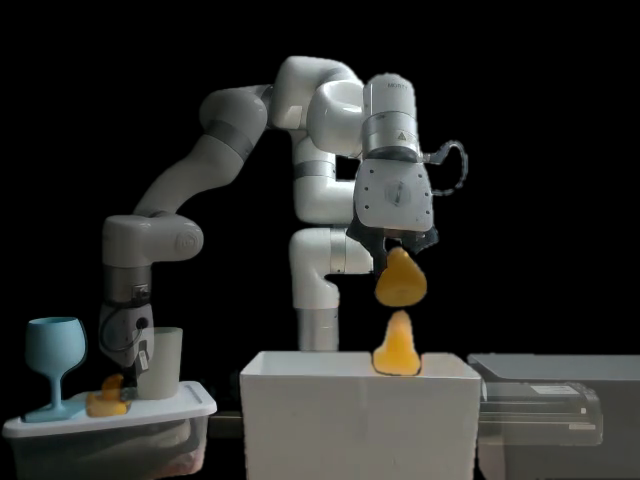} &
        \roundpic{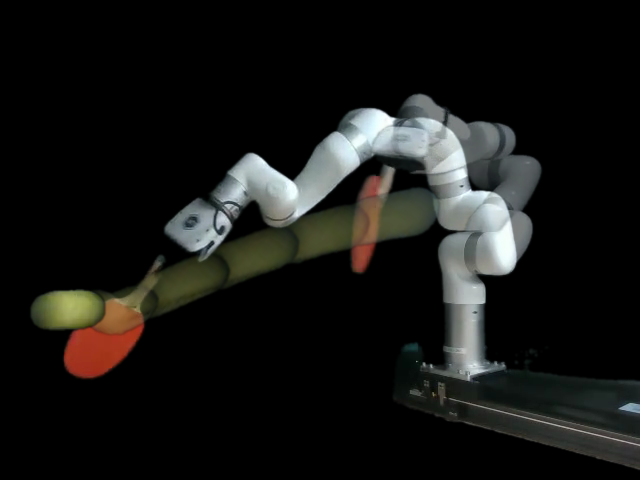} &
        \roundpic{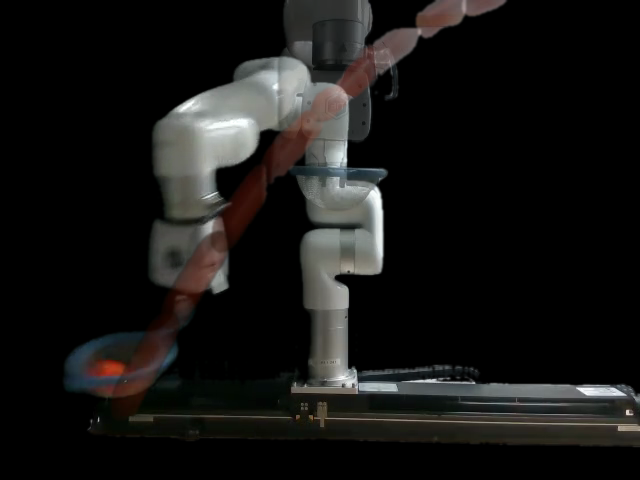} \\
    \end{tabular}
    \caption{
    \methodAbbr{} on four real-world dynamic manipulation tasks. Translucent overlays show object trajectories and transient robot positions; solid silhouettes show the final manipulation pose.
}
    \label{fig:hero}
\end{figure}

%% file: sections/abstract.tex
Vision-Language-Action (VLA) models generalize across static manipulation but fail when objects move during task execution. Policies that act on the current observation assume the scene is stationary until execution, so at any non-trivial object speed the resulting latency exceeds the time available to grasp.
We close this gap with \methodAbbr{} (\methodName{}), a predict-then-act wrapper that augments a frozen VLA with a motion-aware latent world model. A small world model trained on manipulation video forecasts future patch tokens in the VLA's feature space, conditioned on per-token velocity and acceleration from optical flow. A language-and-motion saliency mask concentrates prediction on task-relevant patches, and the model rolls forward for an adaptive horizon, halting when prediction uncertainty crosses a threshold. The frozen action decoder then receives the predicted future tokens in place of the current ones.
\methodAbbr{} adds $4.9$M parameters to a frozen $7$B OpenVLA and reaches $79$ to $97$\% success across $20$ dynamic simulation scenarios where the strongest baseline reaches $31$ to $58$\%. On a physical UFactory xArm~7, \methodAbbr{} succeeds on $29/30$ to $30/30$ on three conveyor and rolling-ball tasks, $23/30$ on paddle interception, and $19/30$ on projectile catching where every baseline scores $0/30$.

%% file: sections/section-1-introduction.tex
\section{Introduction}
\label{sec:introduction}

\begin{leftbarquote}
\itshape ``I skate to where the puck is going to be, not where it has been.'' --- Wayne Gretzky
\end{leftbarquote}

Vision-Language-Action (VLA) models~\citep{Kim2024OpenVLA, OctoModelTeam2024, Brohan2023RT2, Black2024Pi0} have brought general-purpose manipulation policies within reach, but they remain confined to scenes where the world holds still. The moment objects move, on conveyors, in handovers, after release, or under gravity, current VLAs fail at a regime humans handle through anticipatory internal models that track object velocity, plan toward predicted intercept points, and update from contact feedback~\citep{Wolpert1998MOSAIC, Shadmehr1994ForceField}. Policies that act on the current observation assume the scene is stationary until execution, so at any non-trivial object speed the resulting latency exceeds the time available to grasp.
Prior work has pursued two directions to close this gap. Reactive policies~\citep{Zhao2023ACT, Chi2023DiffusionPolicy} dual-system VLAs~\citep{Zhang2024HiRT, Bu2024RoboDual}, and recent concurrent work~\citep{Xie2026DynamicVLA, Tang2025VLASH} shorten the perception-to-action loop through faster control and architectural changes. This shortens the stale window without removing it, and as object speed grows, the residual latency consumes a larger fraction of the reaction window regardless of how fast the policy runs. World model approaches~\citep{Hafner2023DreamerV3, Yang2024Genie, Cen2025WorldVLA, Huang2025LaDiWM} learn forward dynamics and plan through imagined rollouts, but existing formulations roll the entire scene over a fixed horizon and operate either in expensive pixel space or require joint retraining. Neither line addresses which parts of the scene need prediction, how far ahead, or at what latency.
We introduce \methodAbbr{} (\methodName{}), a predict-then-act wrapper that augments a frozen VLA with a motion-aware latent world model in the VLA's feature space. The world model rolls forward only the patches identified as task-relevant by language conditioning and as independently moving by per-patch optical flow, halts each rollout when prediction uncertainty crosses a threshold, and feeds the predicted future tokens to the frozen action decoder. We evaluate \methodAbbr{} on $20$ dynamic simulation scenarios and five physical xArm~7 tasks. In simulation, \methodAbbr{} reaches $79$ to $97$\% success against $31$ to $58$\% for the strongest baseline. On the physical robot, it succeeds on $19/30$ projectile catches with a launcher $2$~m from the robot, where every baseline scores $0/30$. This paper contributes:
\begin{itemize}[leftmargin=*] 
    \item A predict-then-act wrapper for frozen VLAs that enables dynamic-object manipulation without retraining the underlying VLA.
    \item Adaptive compute allocation along both spatial and temporal axes that keeps prediction inside the reactive latency budget, where language-and-motion saliency selects which patches to predict and uncertainty-driven halting determines how far to roll forward.
    \item Explicit kinematic conditioning that propagates per-token velocity and acceleration analytically across rollout steps, extending the world model from constant-velocity to acceleration regimes without learning second-order physics from data.
\end{itemize}

%% file: sections/section-2-related-work.tex
\section{Related Work}
\label{sec:related-work}

\paragraph{Vision-Language-Action Models.}
VLAs unify visual perception, language understanding, and motor control in a single end-to-end architecture, from RT-2's~\citep{Brohan2023RT2} co-fine-tuning of vision-language models with robotic trajectories, to OpenVLA's~\citep{Kim2024OpenVLA} open-source 7B model, Octo's~\citep{OctoModelTeam2024} cross-embodiment transformer, $\pi_0$'s~\citep{Black2024Pi0} flow-matching action head, and GR00T~N1's~\citep{Bjorck2025GR00T} dual-system humanoid controller. These methods act on the current observation with no mechanism for anticipating scene evolution. Dual-system designs such as HiRT~\citep{Zhang2024HiRT} and RoboDual~\citep{Bu2024RoboDual} pair a slow VLM with a fast low-level policy, which shortens the stale window without removing it. \methodAbbr{} addresses this gap by wrapping a frozen VLA with a world model that predicts task-relevant future states.

\paragraph{World Models in Latent Space.}
World models support planning through imagined rollouts. The Dreamer family~\citep{Hafner2020DreamerV1, Hafner2021DreamerV2, Hafner2023DreamerV3} optimizes policies entirely in imagination but requires joint training of policy and dynamics, which is incompatible with a frozen pretrained VLA. TD-MPC2~\citep{Hansen2024TDMPC2} learns latent dynamics for online MPC, with cost scaling in the number of sampled trajectories. Generative video models~\citep{Yang2024Genie, Yang2024UniSim} and autoregressive variants integrated into VLAs~\citep{Cen2025WorldVLA} produce rollouts whose inference exceeds the reactive latency budget. The closest precursor is LaDi-WM~\citep{Huang2025LaDiWM}, a latent diffusion world model that, like \methodAbbr{}, predicts in pretrained feature space. \methodAbbr{} differs by replacing latent diffusion with conditional flow matching, conditioning on explicit per-token kinematic state from optical flow rather than asking the model to infer motion, and rolling forward selectively with adaptive horizon halting where LaDi-WM uses the full state at a fixed horizon. To our knowledge, \methodAbbr{} is the first latent robotic world model to allocate compute adaptively along both spatial and temporal axes.

\paragraph{Model-Predictive Control with Learned Models.}
Sampling-based MPC with learned dynamics~\citep{Williams2017MPPI, Nagabandi2018NeuralMPC, Ebert2018Foresight} evaluates $K$ candidate trajectories per timestep and selects or weights them by predicted return; inference cost scales with $K$. Classical MPC~\citep{Rawlings2017MPC} solves a single optimization over a fixed horizon but requires a differentiable model that is hard to combine with learned VLA action heads. \methodAbbr{} differs from both. The world model rolls forward once to a single predicted future state which feeds directly to the action decoder, with $S{=}5$ samples used for uncertainty estimation rather than action selection.

\paragraph{Reactive and Predictive Dynamic Manipulation.}
Reactive policies~\citep{Zhao2023ACT, Chi2023DiffusionPolicy} reduce observation staleness through action chunking but react to where an object is rather than where it will be; classical interception methods~\citep{Morrison2018Closing, Kopicki2019DynGrasp} rely on hand-designed state representations without large-scale pretraining. Concurrent work approaches dynamic VLAs through faster inference (DynamicVLA~\citep{Xie2026DynamicVLA}, VLASH~\citep{Tang2025VLASH}) or larger world models for offline evaluation (Ctrl-World~\citep{Guo2025CtrlWorld}, VLAW~\citep{Guo2026VLAW}); both accelerate the loop rather than predicting ahead. \methodAbbr{} explicitly predicts where the object will be and acts on that prediction, decoupling effective planning horizon from VLA perception-execution latency.

\paragraph{Video Pretraining for World Models.}
Large-scale video offers object motion dynamics that bootstrap world models without robot demonstration collection~\citep{Cheang2024GR2, Assran2025VJEPA2, Nair2023R3M}. A natural candidate is Ego4D~\citep{Grauman2022Ego4D}, whose 3{,}600+ hours of egocentric video are dominated by outdoor walking and whole-body dynamics that differ from fixed-camera tabletop manipulation. We instead draw from a curated manipulation-focused corpus combining EPIC-Kitchens~\citep{Damen2022EPIC}, Something-Something~V2~\citep{Goyal2017SSv2}, DROID~\citep{DROID2024}, and Bridge~V2~\citep{Walke2023BridgeV2}, whose flow statistics match the deployment setting. Because \methodAbbr{}'s world model operates in VLA feature space rather than pixel space, any manipulation-relevant motion video provides useful training signal.

%% file: sections/section-3-problem.tex
\section{Problem Statement}
\label{sec:problem}

\paragraph{Setting.}
Consider a robot operating in an environment where objects may move independently of the robot's actions. At each time step $t$, the robot receives a monocular RGB observation $o_t$ and a natural-language task instruction $\ell$. A frozen pretrained VLA provides a vision encoder $\phi$ that maps observations to $N$ patch tokens $v_t = \phi(o_t) \in \mathbb{R}^{N \times d}$, where $d$ is the patch token dimension. The VLA also provides a language encoder that produces instruction embeddings $e_\ell$, and an action decoder $\pi$ that generates a motor command $a_t = \pi(v_t, e_\ell) \in \mathcal{A}$. The scene state evolves from two sources, robot-induced change from $a_t$ and independent change from external agents or physical processes. Policies that act on the current observation implicitly assume the scene remains static until execution, so when independent change is present, actions planned at time $t$ target an outdated scene state.

\paragraph{Objective.}
We aim to augment the frozen VLA with the ability to anticipate independent scene change, so that the action decoder receives an estimate of future scene state in place of the stale current observation. The augmentation should preserve the frozen VLA's pretrained capabilities and operate within the real-time control loop, leaving the underlying action decoder and language grounding unchanged. 

\paragraph{Desiderata.}
An effective predict-then-act loop must satisfy two desiderata. First, \textbf{scene-adaptive prediction horizon}. The realized horizon should match per-scene predictability rather than being fixed. Linear, low-uncertainty motion admits long horizons that anticipate contact well in advance, whereas chaotic or post-collision motion admits only short horizons before compounding prediction error dominates, so a solution must expose some signal of per-scene prediction reliability. Second, \textbf{real-time execution}. The full loop from observation to predicted future to action must complete within a latency budget $\Delta t_{\max}$, since predictions arriving after $\Delta t_{\max}$ are stale by the time they reach the action decoder. %

\paragraph{Scope.}
This work targets \emph{dynamic scenes}, settings in which objects move independently of the robot during task execution, including human-initiated object displacement, conveyor transport, and post-contact rolling.  The method assumes object motion is well-approximated by constant acceleration (including the zero-acceleration limit of constant velocity) over the prediction horizon.  Highly stochastic post-collision dynamics violate this assumption; we evaluate such scenarios as a stress test and report graceful degradation rather than competitive performance (Appendix~\ref{app:results-stress}).

%% file: sections/section-4-method.tex
\section{Method}
\label{sec:method}

\input{figures/architecture}

\methodAbbr{} augments a frozen VLA with a motion-aware latent world model so that the robot acts on predicted future states rather than stale observations. At each time step $t$, the frozen VLA vision encoder $\phi$ maps the current RGB observation $o_t$ to $N$ patch tokens $v_t \in \mathbb{R}^{N \times d}$. Standard VLAs feed $v_t$ directly to the action decoder and act on what the world looked like at time $t$, which is already stale by the time the action executes. \methodAbbr{} instead predicts $\hat{v}_{t+K}$, the patch tokens corresponding to the scene state $K$ steps in the future, and feeds those predicted tokens to the unchanged action decoder (Figure~\ref{fig:architecture}).
Three observations shape the pipeline. First, most patches in a manipulation scene do not move and do not warrant compute; only patches with task-relevant motion need prediction. Second, the right prediction horizon $K$ depends on the scene; linear motion admits long lookahead, chaotic motion does not. Third, the entire prediction must complete within a reactive manipulation latency budget on the order of $200$~ms. These observations motivate spatial compute allocation (Section~\ref{sec:method-spatial}), temporal compute allocation (Section~\ref{sec:method-allocator}), and the dynamics model choice (Section~\ref{sec:method-world-model}). Section~\ref{sec:method-pretraining} describes training. Appendix~\ref{app:method} gives full architectural details, ablation rationale, and hyperparameters.

\subsection{Language-and-Motion Saliency}
\label{sec:method-spatial}

RAFT~\citep{Teed2020RAFT} computes optical flow between three consecutive observations, which patch-level pooling and finite-differencing convert to per-token velocity $V$ and acceleration $A$ (Appendix~\ref{app:method-motion}). We concatenate each patch token with its motion descriptor to form motion-enriched tokens $\tilde{v}_t \in \mathbb{R}^{N \times (d+4)}$. To allocate compute spatially, \methodAbbr{} predicts only on tokens that are either language-relevant or independently moving. A learned cross-attention layer attends from $\tilde{v}_t$ to the instruction embeddings $e_\ell$, producing per-token language relevance scores $\widetilde{M} \in [0,1]^N$. Tokens whose velocity magnitude exceeds a calibrated threshold $\tau_{\text{flow}}$ receive an elevated relevance floor,
\begin{equation}
\label{eq:motion-saliency}
    M_i = \max\!\bigl(\widetilde{M}_i,\;
    \alpha_{\text{motion}} \cdot
    \mathbf{1}[\|V_i\| > \tau_{\text{flow}}]\bigr).
\end{equation}
\methodAbbr{} forms the selected index set $\mathcal{S} = \{i : M_i > 0.2\}$, typically yielding $30$ to $60$ tokens out of $N=196$ in our experiments. Appendix~\ref{app:method-spatial} gives morphological dilation, threshold calibration, and $\alpha_{\text{motion}}$.

\subsection{Flow-Matching World Model with Kinematic Conditioning}
\label{sec:method-world-model}

The world model operates in VLA feature space (rather than pixels) to align with the action decoder's input distribution and avoid expensive reconstruction. \methodAbbr{} uses conditional flow matching~\citep{Lipman2023FlowMatching}, which produces high-quality samples in a few Euler steps~\citep{Esser2024SD3} against $100$+ for diffusion, keeping the world model within the per-step latency budget (Appendix~\ref{app:method-latency}). Drawing $S$ samples in parallel yields both a mean prediction and a sample-variance uncertainty estimate at no extra cost. A 4-layer transformer encoder compresses the $|\mathcal{S}|$ saliency-selected, motion-enriched tokens $\tilde{v}_t^{(\mathcal{S})}$ into a compact latent $z_t$, and the dynamics model rolls forward autoregressively,
\begin{equation}
\label{eq:dynamics}
    z_{t+k} \sim p\!\left(z_{t+k} \,\middle|\, z_{t+k-1},\; V_k^{(\mathcal{S})},\; A^{(\mathcal{S})}\right),
    \quad k = 1, \ldots, K.
\end{equation}
To avoid asking the world model to learn full physics from data, \methodAbbr{} propagates the velocity conditioning analytically across rollout steps under constant-acceleration kinematics,
\begin{equation}
\label{eq:kinematic-update}
    V_k = V_0 + A \cdot k \cdot \Delta t,
\end{equation}
where $V_0$ is the initial RAFT-derived velocity and $A$ the finite-differenced acceleration. This captures the dominant trajectory curvature in acceleration-driven scenarios such as gravitational pull and friction decay. An MLP decoder reconstructs predicted patch tokens, which are spliced back into the full grid (unselected patches pass through unchanged) before being passed to the frozen action decoder, which produces $a_t = \pi(\hat{v}_{t+K}, e_\ell)$. The spliced grid is a functional surrogate for the future observation rather than a visual reconstruction (Appendix~\ref{app:method-architecture}). Appendix~\ref{app:method-architecture} details architecture dimensions; Appendix~\ref{app:method-alternatives} compares against diffusion and dropout-based GRU alternatives.

\subsection{Adaptive Horizon Halting}
\label{sec:method-allocator}

To allocate compute temporally, \methodAbbr{} halts each rollout when prediction uncertainty crosses a threshold. The $S$ trajectory samples drawn at each step (Section~\ref{sec:method-world-model}) provide a per-step scene-level uncertainty estimate as their mean per-token variance,
\begin{equation}
\label{eq:variance}
    \bar{u}_{t+k} = \frac{1}{|\mathcal{S}|} \sum_{i \in \mathcal{S}}
    \frac{1}{S} \sum_{j=1}^{S}
    \bigl\| z_{t+k}^{(j)}[i] - \bar{z}_{t+k}[i] \bigr\|^2,
\end{equation}
where $z_{t+k}^{(j)}$ is the $j$-th sample and $\bar{z}_{t+k}$ the sample mean. The rollout halts at the first step $K$ satisfying $\bar{u}_{t+K} > \tau_u$, or at maximum horizon $K_{\max}$ otherwise. Linear, low-uncertainty motion runs longer; chaotic or post-collision motion halts earlier. In experiments we set $K_{\max} = 10$ and $\tau_u$ to the 90th percentile of $\bar{u}$ on the training set.

\subsection{Training}
\label{sec:method-pretraining}

The world model is trained in two phases. Pretraining on a corpus of manipulation video establishes a prior over object dynamics in VLA feature space, since any video containing manipulation-relevant motion provides useful training signal once features are extracted by $\phi$. Fine-tuning on $\approx 200$ in-domain xArm~7 trajectories closes the domain gap to the deployment setting. The dynamics model uses the conditional flow matching objective; the encoder and decoder use MSE reconstruction over the rollout horizon. The feature alignment layer is trained separately. Appendix~\ref{app:method-training} specifies the pretraining corpus, curriculum, loss functions, and hyperparameters.

%% file: figures/architecture.tex
\begin{figure}[t]
    \centering
    \includegraphics[width=\linewidth]{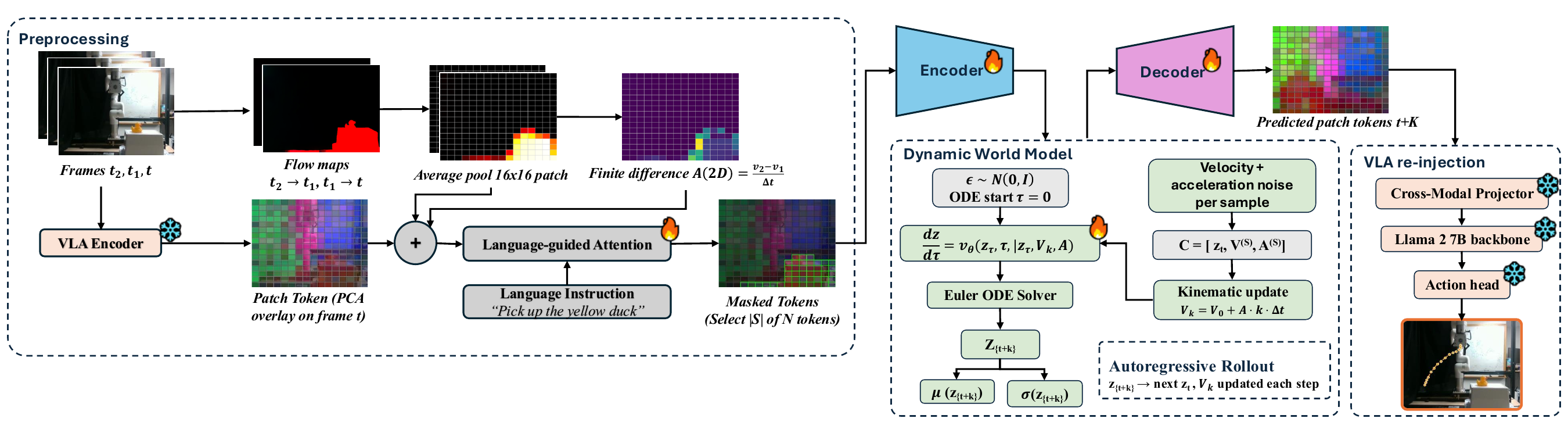}
    \caption{
    \textbf{\methodAbbr{} architecture.}
    RAFT optical flow~\citep{Teed2020RAFT} provides per-patch velocity and acceleration. A Language-and-motion saliency mask selects a task-relevant subset $\mathcal{S}$ of $N$ patch tokens, which are encoded, rolled forward by a conditional flow-matching world model with analytical kinematic conditioning $V_k = V_0 + A \cdot k \cdot \Delta t$, decoded, and spliced back into the full token grid for the frozen action decoder. Flame icons mark trained components ($\sim$$4.9$M parameters total); snowflakes mark components frozen from OpenVLA. Figure~\ref{fig:architecture-full} in Appendix~\ref{app:method-architecture-figure} shows the full breakdown including autoregressive rollout, adaptive halting, and feature alignment.
    }
    \label{fig:architecture}
\end{figure}

%% file: sections/section-5-experiments.tex
\section{Experiments}
\label{sec:experiments}

We evaluate \methodAbbr{} in simulation across $20$ dynamic-object manipulation scenarios and on a physical UFactory xArm~7 across five dynamic tasks. Ablations isolate the contribution of each component.

\subsection{Experimental Setup}
\label{sec:exp-setup}

\paragraph{Simulation.}
We use custom MuJoCo~\citep{Todorov2012MuJoCo} environments built around a Franka Emika Panda~\citep{Haddadin2022Franka} arm spanning four motion categories: constant-velocity transport (conveyor, beam, pole push with red cup), gravity-driven motion (rolling ball), reactive contact (air hockey, ballistic catching), and chaotic post-collision dynamics (multi-deflection pinball, occlusion deflection, plinko drop). Each scenario is procedurally generated with randomized object positions, motion parameters, and lighting. Appendix~\ref{app:setup-sim} gives scene specifications and per-scenario success criteria.

\paragraph{Physical platform.}
A UFactory xArm~7 with a fixed third-person RGB camera at $224 \times 224$ and a parallel-jaw gripper, running on a single NVIDIA RTX~4090 at 171~ms per action step, evaluated on five tasks: (i) static object onto moving conveyor box, (ii) static box receiving moving object, (iii) projectile interception with a ping-pong paddle, (iv) stopping a rolling ball, and (v) catching a projectile from a mechanical launcher. A speed-sensitivity sweep varies conveyor speed from $0$ to $25$~cm/s. Appendix~\ref{app:setup-physical} gives hardware details.

\paragraph{Baselines.}
Six learned baselines span the relevant axes of dynamic manipulation: Open-loop VLA and Retargeting VLA (closed-loop)~\citep{Kim2024OpenVLA}; VLA + Fast Replan~\citep{Kim2025OpenVLAOFT} with action chunking; Realtime ACT~\citep{Zhao2023ACT} and Streaming Diffusion Policy~\citep{Chi2023DiffusionPolicy} for chunked real-time execution; and DreamVLA~\citep{Zhang2025DreamVLA}, a world-model-based VLA that predicts dynamic regions and depth before generating actions. Two motion-informed baselines feed the same frozen OpenVLA. Track-and-Intercept is a classical tracker that extrapolates the object centroid under a constant-acceleration fit over \methodAbbr{}'s realized horizon and warps the frame so the object appears at its predicted location, a pixel-space analogue of our token splice. LaDi-WM~\citep{Huang2025LaDiWM} is retrained in OpenVLA feature space. Appendix~\ref{app:setup-baselines} gives per-baseline hyperparameters and training data.

\paragraph{Protocol and configuration.}
In simulation, each cell reports mean $\pm$ standard error over $5$ training seeds with $100$ rollouts per seed ($500$ rollouts per cell). On the physical robot, each task reports successes out of $30$ attempts ($10$ per speed point for the speed sweep). All \methodAbbr{} results use: frozen OpenVLA, RAFT~\citep{Teed2020RAFT} optical flow, kinematic and acceleration-conditioned velocity model, language-and-motion saliency masking, $4$-layer encoder, $S = 5$ flow-matching samples with $5$ Euler steps, $K_{\max} = 10$, and $\tau_u$ at the $90$th percentile of $\bar{u}$ on the training set. Realized horizon averages $3$ to $5$ steps. Full hyperparameters in Table~\ref{tab:hyperparams}.

\subsection{Simulation Results}
\label{sec:exp-sim}

We evaluate three aspects of dynamic-object manipulation in simulation: success across constant-velocity and acceleration regimes (Table~\ref{tab:results-dynamic}), robustness to object speed (Table~\ref{tab:speed}), and generalization to complex scenarios with multi-object scenes, occlusion, and mid-flight trajectory changes (Table~\ref{tab:results-complex}).

\input{tables/results-dynamic}

\paragraph{Constant velocity and acceleration.}
Table~\ref{tab:results-dynamic} shows \methodAbbr{} outperforming all baselines on every scenario in both regimes, with the largest gap on acceleration-regime scenarios where the strongest baseline (DreamVLA) reaches $30$ to $48\%$ while \methodAbbr{} stays above $87\%$. Performance is also stable across regimes (within $10$ percentage points), while baselines degrade substantially. Inspecting failure modes, baselines without explicit prediction (Retargeting VLA, VLA~+~Fast~Replan, Realtime ACT) most often fail by grasping where the object was rather than where it will be; the gripper closes on empty space behind the moving object. The explicit kinematic update (Eq.~\ref{eq:kinematic-update}) lets \methodAbbr{} target the predicted intercept point instead.

\input{tables/results-speed}

\paragraph{Robustness to object speed.}
Table~\ref{tab:speed} shows \methodAbbr{} maintaining over $95\%$ success across the $0$ to $40$~cm/s sweep while every baseline degrades steadily. The static case ($0$~cm/s) shows that all methods are well-matched on perception and grasping in isolation; the gap opens once independent object motion is introduced, suggesting the gap is attributable to predictive lookahead rather than confounding perception or action-decoder differences. The $2.2$-point residual at $40$~cm/s comes from RAFT error at the high-velocity end of the flow regime; the motion-estimator ablation (Appendix~\ref{app:abl-motion}) supports this.

\input{tables/results-complex}

\paragraph{Complex scenarios.}
Table~\ref{tab:results-complex} shows \methodAbbr{} reaching $79$ to $96\%$ on every complex scenario where the strongest baseline reaches at most $60\%$. The Occlusion (Occ) column is the cleanest demonstration that prediction is qualitatively distinct from reactivity. Every baseline except Realtime ACT scores exactly $0\%$ on Occ, because catching an object that has disappeared behind an occluder requires predicting where it will reappear, which a closed-loop policy looking only at the latest frame cannot solve. \methodAbbr{} reaches $79.4\%$.

\paragraph{Motion-informed baselines.}
Track-and-Intercept averages 86.8\% on constant-velocity scenes against $93.5\%$ for \methodAbbr{}, so a kinematic prior alone does much of the work when motion is clean. It reaches 51.2\% on Ballistic Catch against $87.4\%$, as a 2D warp cannot represent scale change and leaves disocclusion holes, and 33.6\% on L+MO (language + multiple-objects) against $95.6\%$, as the tracker cannot resolve the instruction referent. LaDi-WM averages 63.0\% on constant-velocity scenes and 18.6\% on Occlusion, as full-state prediction at a fixed horizon must infer motion from features alone.

\subsection{Physical Robot Experiments}
\label{sec:exp-physical}

\input{tables/results-physical}

Table~\ref{tab:physical} shows \methodAbbr{} as the strongest method on every physical task, with $30/30$ on conveyor pick-and-place, $30/30$ on rolling ball, and $19/30$ on projectile catching where every baseline scores $0/30$. Transfer is tight on smooth-motion tasks ($97.3\%$ sim, $30/30$ real on conveyor) but degrades on contact-physics tasks (paddle hitting $94.6\%$ sim, $23/30$ real; projectile catching $87.4\%$ sim, $19/30$ real), reflecting limits of MuJoCo's collision model, gripper compliance, and camera shutter behavior. The $11$ unsuccessful projectile catches break down as follows. Five came from joint-limit guards halting the arm short of an otherwise geometrically-correct intercept, indicating the predicted target was reachable in principle but not from the current configuration. The remaining six split between the end-effector arriving just after the ball passed and the ball bouncing out of the net on contact. The speed-sensitivity sweep (right half of Table~\ref{tab:physical}) shows the speed-robustness story transferring to hardware. \methodAbbr{} is the only method exceeding chance performance beyond $5$~cm/s.

\subsection{Ablation Studies}
\label{sec:exp-ablations}

Table~\ref{tab:ablations-summary} reports mean success rate for each ablated configuration across the four scenarios in the relevant scene group (constant-velocity for motion estimator, spatial masking, and fixed horizon; acceleration/deceleration for velocity model). Per-scenario tables and the world model architecture ablation (samples $S$, encoder depth) are in Appendix~\ref{app:ablations}.

\input{tables/ablations-summary}

\paragraph{What each component buys.}
Explicit motion and the analytic kinematic update drive accuracy. Feeding the world model the same three frames' VLA tokens with no flow input and matched capacity reaches 74.2\% against $93.5\%$, since sub-patch displacements are a noisy motion signal in pooled semantic tokens. Saliency masking and adaptive halting are accuracy-neutral on constant-velocity scenes, where full prediction and fixed $K{=}5$ match the default within error. Their contribution is fitting the latency budget and extending the method to chaotic motion. Predicting all $196$ tokens raises latency to 250~ms, past the $200$~ms budget (Table~\ref{tab:abl-masking}), and on Plinko at the $1600$~ms window adaptive halting reaches $48.6\%$ against 31.4\% for fixed $K{=}5$ (Table~\ref{tab:abl-horizon-stress}), as post-collision variance truncates rollouts at the peg contact. Removing the alignment layer lowers Ballistic Catch to 61.1\% (Table~\ref{tab:abl-splice}), still above every baseline in Table~\ref{tab:results-complex}, so the $0.1$M layer recovers fidelity rather than supplying the policy.

%% file: tables/results-dynamic.tex
\begin{table}[t]
    \centering
    \small
    \caption{Task success rate (\%) on dynamic scenes across constant-velocity and acceleration/deceleration regimes. The two rows below the first rule are motion-informed baselines that feed the same frozen OpenVLA. Mean $\pm$ standard error over 5 seeds with 100 rollouts each.}
    \label{tab:results-dynamic}
    \setlength{\tabcolsep}{3.5pt}
    \resizebox{\linewidth}{!}{%
    \begin{tabular}{@{}lcccc|cccc@{}}
        \toprule
        & \multicolumn{4}{c|}{\textbf{Constant velocity}}
        & \multicolumn{4}{c}{\textbf{Acceleration / deceleration}} \\
        \cmidrule(lr){2-5} \cmidrule(lr){6-9}
        \textbf{Method}
            & \textbf{Conv.} & \textbf{Beam} & \textbf{Pole} & \textbf{Roll}
            & \textbf{Conv.} & \textbf{Beam} & \textbf{Pole} & \textbf{Roll (g.)} \\
        \midrule
        Open-loop VLA              & $0.0 \pm 0.0$ & $0.0 \pm 0.0$ & $0.0 \pm 0.0$ & $0.0 \pm 0.0$
                                    & $0.0 \pm 0.0$ & $0.4 \pm 0.6$ & $0.0 \pm 0.0$ & $0.0 \pm 0.0$ \\
        Retargeting VLA            & $13.7 \pm 3.8$ & $22.3 \pm 4.0$ & $13.3 \pm 3.2$ & $7.7 \pm 2.9$
                                    & $1.0 \pm 1.4$ & $10.7 \pm 3.1$ & $1.0 \pm 1.4$ & $0.6 \pm 1.1$ \\
        VLA + Fast Replan          & $24.0 \pm 4.6$ & $32.7 \pm 4.2$ & $22.3 \pm 3.7$ & $16.7 \pm 4.0$
                                    & $22.3 \pm 4.2$ & $31.3 \pm 4.6$ & $21.7 \pm 4.1$ & $16.3 \pm 3.7$ \\
        Realtime ACT               & $51.7 \pm 5.1$ & $60.3 \pm 4.7$ & $48.7 \pm 4.9$ & $37.0 \pm 5.3$
                                    & $31.7 \pm 4.6$ & $41.0 \pm 4.9$ & $40.3 \pm 4.9$ & $31.0 \pm 4.6$ \\
        Streaming DP               & $40.7 \pm 5.2$ & $50.3 \pm 4.8$ & $39.3 \pm 4.5$ & $28.7 \pm 4.6$
                                    & $21.0 \pm 4.1$ & $29.7 \pm 4.6$ & $29.0 \pm 4.5$ & $19.7 \pm 4.0$ \\
        DreamVLA                 & $58.3 \pm 4.7$ & $47.7 \pm 5.1$ & $57.3 \pm 5.0$ & $46.0 \pm 5.2$
                                    & $40.3 \pm 4.9$ & $38.7 \pm 4.9$ & $48.0 \pm 5.0$ & $30.7 \pm 4.6$ \\
        \midrule
        Track-and-Intercept VLA    & $91.2 \pm 2.6$ & $85.4 \pm 3.3$ & $88.6 \pm 3.0$ & $82.0 \pm 3.7$
                                    & $71.4 \pm 4.3$ & $74.8 \pm 4.1$ & $69.2 \pm 4.4$ & $63.6 \pm 4.6$\\
        LaDi-WM (OpenVLA feat.)    & $68.4 \pm 4.4$ & $61.2 \pm 4.7$ & $66.8 \pm 4.5$ & $55.4 \pm 4.9$
                                    & $49.6 \pm 4.8$ & $47.2 \pm 4.9$ & $55.8 \pm 4.7$ & $38.4 \pm 4.6$ \\
        \midrule
        \textbf{\methodAbbr{} (ours)}
            & $\mathbf{97.3 \pm 1.5}$ & $\mathbf{91.7 \pm 2.8}$ & $\mathbf{95.0 \pm 2.3}$ & $\mathbf{90.0 \pm 3.1}$
            & $\mathbf{87.7 \pm 3.3}$ & $\mathbf{91.0 \pm 3.9}$ & $\mathbf{93.7 \pm 6.1}$ & $\mathbf{89.0 \pm 0.9}$ \\
        \bottomrule
    \end{tabular}%
    }
\end{table}

%% file: tables/results-speed.tex
\begin{table}[t]
    \centering
    \small
    \caption{Task success rate (\%) versus conveyor speed on the conveyor + red cup scenario. Mean $\pm$ standard error over 5 seeds with 100 rollouts each.}
    \label{tab:speed}
    \begin{tabular}{@{}lccccc@{}}
        \toprule
        \textbf{Method}
            & \textbf{0 cm/s}
            & \textbf{10 cm/s}
            & \textbf{20 cm/s}
            & \textbf{30 cm/s}
            & \textbf{40 cm/s} \\
        \midrule
        Open-loop VLA      & $96.4 \pm 1.9$ & $0.0 \pm 0.0$  & $0.0 \pm 0.0$  & $0.0 \pm 0.0$  & $0.0 \pm 0.0$ \\
        Retargeting VLA    & $96.8 \pm 1.8$ & $40.6 \pm 4.9$ & $31.2 \pm 4.4$ & $22.4 \pm 4.1$ & $14.6 \pm 3.4$ \\
        Realtime ACT       & $97.2 \pm 1.6$ & $78.4 \pm 3.9$ & $64.8 \pm 4.4$ & $51.4 \pm 4.6$ & $41.0 \pm 4.7$ \\
        Streaming Diffusion Policy   & $95.4 \pm 2.1$ & $31.6 \pm 4.4$ & $27.2 \pm 4.0$ & $22.4 \pm 4.1$ & $18.6 \pm 3.7$ \\
        DreamVLA         & $96.8 \pm 1.8$ & $79.2 \pm 3.8$ & $70.4 \pm 4.3$ & $60.2 \pm 4.4$ & $47.2 \pm 4.1$ \\
        \textbf{\methodAbbr{} (ours)}
                           & $\mathbf{97.6 \pm 1.5}$
                           & $\mathbf{97.4 \pm 1.7}$
                           & $\mathbf{97.0 \pm 1.8}$
                           & $\mathbf{96.6 \pm 2.0}$
                           & $\mathbf{95.4 \pm 2.2}$ \\
        \bottomrule
    \end{tabular}
\end{table}

%% file: tables/results-complex.tex
\begin{table}[t]
    \centering
    \small
    \caption{Task success rate (\%) on eight complex dynamic scenarios (Air Hockey, Ballistic Catch, Multi-Object, Place in Box, Multi-Balls, Occlusion, Mid-flight deflection, Language + Multi-Object). Full scenario descriptions in Appendix~\ref{app:setup-sim}. Mean $\pm$ standard error over $5$ seeds with $100$ rollouts each.}
    \label{tab:results-complex}
    \resizebox{\linewidth}{!}{%
    \begin{tabular}{@{}lcccccccc@{}}
        \toprule
        \textbf{Method}
            & \textbf{AH} & \textbf{BC} & \textbf{MO} & \textbf{PiB}
            & \textbf{MB} & \textbf{Occ} & \textbf{Mid} & \textbf{L+MO} \\
        \midrule
        Open-loop VLA      & $0.0 \pm 0.0$  & $0.0 \pm 0.0$  & $0.0 \pm 0.0$  & $0.0 \pm 0.0$
                            & $0.0 \pm 0.0$  & $0.0 \pm 0.0$  & $0.0 \pm 0.0$  & $0.0 \pm 0.0$ \\
        Retargeting VLA    & $10.6 \pm 3.0$ & $0.0 \pm 0.0$  & $0.0 \pm 0.0$  & $0.0 \pm 0.0$
                            & $0.0 \pm 0.0$  & $0.0 \pm 0.0$  & $0.0 \pm 0.0$  & $0.0 \pm 0.0$ \\
        VLA + Fast Replan  & $0.0 \pm 0.0$  & $0.0 \pm 0.0$  & $31.6 \pm 4.4$ & $22.4 \pm 4.1$
                            & $0.0 \pm 0.0$  & $0.0 \pm 0.0$  & $0.0 \pm 0.0$  & $0.0 \pm 0.0$ \\
        Realtime ACT       & $31.2 \pm 4.4$ & $22.4 \pm 4.1$ & $40.6 \pm 4.9$ & $41.2 \pm 4.8$
                            & $22.4 \pm 4.1$ & $11.4 \pm 3.0$ & $60.2 \pm 4.4$ & $31.6 \pm 4.4$ \\
        Streaming DP       & $11.4 \pm 3.0$ & $41.6 \pm 4.8$ & $41.4 \pm 4.6$ & $22.4 \pm 4.1$
                            & $0.0 \pm 0.0$  & $0.0 \pm 0.0$  & $41.6 \pm 4.8$ & $30.4 \pm 4.3$ \\
        DreamVLA         & $31.2 \pm 4.4$ & $49.1 \pm 9.1$ & $43.5 \pm 4.5$ & $41.2 \pm 4.8$
                            & $40.6 \pm 4.9$ & $0.0 \pm 0.0$  & $49.4 \pm 4.7$ & $45.2 \pm 4.8$ \\
        \midrule
        Track-and-Intercept VLA & $62.4 \pm 3.5$ & $51.2 \pm 4.7$ & $38.6 \pm 1.6$ & $71.8 \pm 6.2$
                            & $29.4 \pm 0.3$ & $34.8 \pm 4.5$ & $27.6 \pm 2.2$ & $33.6 \pm 2.4$ \\
        LaDi-WM (OpenVLA feat.) & $48.2 \pm 9.7$ & $44.6 \pm 1.8$ & $52.4 \pm 1.7$ & $50.8 \pm 3.8$
                            & $41.2 \pm 5.7$ & $18.6 \pm 3.7$ & $36.4 \pm 2.6$ & $49.8 \pm 1.8$ \\
        \midrule
        \textbf{\methodAbbr{} (ours)}
            & $\mathbf{94.6 \pm 2.3}$ & $\mathbf{87.4 \pm 3.2}$ & $\mathbf{95.4 \pm 2.2}$ & $\mathbf{95.8 \pm 2.0}$
            & $\mathbf{87.2 \pm 3.3}$ & $\mathbf{79.4 \pm 3.8}$ & $\mathbf{94.4 \pm 2.4}$ & $\mathbf{95.6 \pm 2.1}$ \\
        \bottomrule
    \end{tabular}%
    }
\end{table}

%% file: tables/results-physical.tex
\begin{table}[t]
    \centering
    \small
    \setlength{\tabcolsep}{3pt}
    \caption{Physical robot experiments on a UFactory xArm~7. \textbf{Left}: five dynamic manipulation tasks (out of $30$ attempts). \textbf{Right}: conveyor speed sensitivity (out of $10$ attempts per speed point).}
    \label{tab:physical}
    \resizebox{\linewidth}{!}{%
    \begin{tabular}{@{}l ccccc | cccccc@{}}
        \toprule
        & \multicolumn{5}{c|}{\textbf{Tasks}}
        & \multicolumn{6}{c}{\textbf{Speed sensitivity (cm/s)}} \\
        \cmidrule(lr){2-6} \cmidrule(lr){7-12}
        \textbf{Method}
            & \shortstack[c]{\textbf{Static obj.,}\\\textbf{mov.\ box}}
            & \shortstack[c]{\textbf{Static box,}\\\textbf{mov.\ obj.}}
            & \shortstack[c]{\textbf{Paddle}\\\textbf{hit}}
            & \shortstack[c]{\textbf{Stop}\\\textbf{ball}}
            & \shortstack[c]{\textbf{Catch}\\\textbf{proj.}}
            & \textbf{0} & \textbf{5} & \textbf{10} & \textbf{15} & \textbf{20} & \textbf{25} \\
        \midrule
        Open-loop VLA              & 0/30  & 0/30  & 0/30  & 0/30  & 0/30  & 10/10 & 0/10  & 0/10  & 0/10  & 0/10  & 0/10 \\
        Retargeting VLA            & 0/30  & 0/30  & 0/30  & 0/30  & 0/30  & 10/10 & 4/10  & 0/10  & 0/10  & 0/10  & 0/10 \\
        VLA + Fast Replan          & 0/30  & 0/30  & 0/30  & 0/30  & 0/30  & --    & --    & --    & --    & --    & -- \\
        Realtime ACT               & 9/30  & 5/30  & 9/30  & 2/30  & 0/30  & 10/10 & 8/10  & 0/10  & 0/10  & 0/10  & 0/10 \\
        Streaming Diffusion Policy & 3/30  & 3/30  & 0/30  & 0/30  & 0/30  & 10/10 & 0/10  & 3/10  & 0/10  & 0/10  & 0/10 \\
        DreamVLA                   & 12/30 & 9/30  & 1/30  & 12/30 & 0/30  & 10/10 & 2/10  & 0/10  & 0/10  & 0/10  & 0/10 \\
        \textbf{\methodAbbr{} (ours)}
            & \textbf{30/30} & \textbf{29/30} & \textbf{23/30} & \textbf{30/30} & \textbf{19/30}
            & \textbf{10/10} & \textbf{10/10} & \textbf{10/10} & \textbf{7/10} & \textbf{6/10} & \textbf{2/10} \\
        \bottomrule
    \end{tabular}%
    }
\end{table}

%% file: tables/ablations-summary.tex
\begin{table}[t]
    \centering
    \footnotesize
    \setlength{\tabcolsep}{3pt}
    \renewcommand{\arraystretch}{0.95}
    \caption{Ablation summary. Mean success rate (\%) $\pm$ standard error across four scenarios in the relevant scene group (constant-velocity for motion estimator, spatial masking, fixed horizon; acceleration/deceleration for velocity model). Full per-scenario tables in Appendix~\ref{app:ablations}.}
    \label{tab:ablations-summary}
    \resizebox{\linewidth}{!}{%
    \begin{tabular}{@{}lc@{\hspace{1em}}lc@{\hspace{1em}}lc@{\hspace{1em}}lc@{}}
        \toprule
        \multicolumn{2}{c}{\textbf{Motion estimator}}
            & \multicolumn{2}{c}{\textbf{Velocity model}}
            & \multicolumn{2}{c}{\textbf{Spatial masking}}
            & \multicolumn{2}{c}{\textbf{Fixed horizon $K$}} \\
        \cmidrule(lr){1-2} \cmidrule(lr){3-4} \cmidrule(lr){5-6} \cmidrule(lr){7-8}
        RAFT (ours)         & $\mathbf{93.5 \pm 1.8}$ & Kinematic + accel (ours) & $\mathbf{90.7 \pm 1.1}$  & Language + motion (ours) & $\mathbf{93.8 \pm 1.2}$ & $K = 5$ (=adaptive) & $\mathbf{93.8 \pm 1.2}$ \\
        CoTracker3          & $84.0 \pm 2.7$          & Kinematic only           & $82.5 \pm 3.6$           & Full prediction          & $91.2 \pm 2.7$          & $K = 8$             & $92.9 \pm 1.1$ \\
        Implicit 3-frame context & $74.2 \pm 3.1$ & Learned dynamics         & $78.7 \pm 7.9$           & Motion only              & $84.0 \pm 1.5$          & $K = 12$            & $86.3 \pm 3.0$ \\
        Farneb\"ack         & $68.9 \pm 4.5$          & Accel-token only         & $78.5 \pm 5.8$           & Language only            & $77.9 \pm 3.9$          & $K = 3$             & $79.5 \pm 1.8$ \\
        Frame differencing  & $38.5 \pm 3.4$          & Constant $V_k = V_0$     & $71.7 \pm 10.4$          & Uniform random ($30\%$)  & $46.4 \pm 2.5$          & $K = 2$             & $60.2 \pm 1.9$ \\
        No velocity input   & $0.0 \pm 0.0$           & Exponential decay        & $65.3 \pm 11.0$          &                          &                          & $K = 1$             & $50.0 \pm 4.0$ \\
        \bottomrule
    \end{tabular}%
    }
\end{table}

%% file: sections/section-6-conclusion.tex
\section{Conclusion}
\label{sec:conclusion}

\methodAbbr{} augments a frozen VLA with a motion-aware latent world model so the robot acts on predicted future states rather than stale observations. A $4.9$M-parameter wrapper around an unmodified $7$B OpenVLA backbone reaches $79$ to $97$\% success across $20$ dynamic simulation scenarios where the strongest baseline reaches $31$ to $58$\%, and succeeds on $30/30$ and $29/30$ on two conveyor tasks, $30/30$ on a rolling-ball task, $23/30$ on paddle interception, and $19/30$ on projectile catching where every baseline fails entirely. The full pipeline runs in $\approx 158$~ms per action step, within the $\approx 200$~ms reactive manipulation budget. Ablations show that explicit RAFT-based motion estimation and the analytical kinematic update drive accuracy, while language-and-motion saliency and adaptive halting keep prediction within the latency budget. The broader contribution is an architectural template for adding predictive lookahead to pretrained VLAs without modifying or retraining the underlying model.

\subsection{Limitations and Future Work}
\label{sec:limitations}

The constant-acceleration kinematic update breaks down under chaotic post-collision dynamics. Plinko-drop performance drops to $48.6\%$ at the longest reaction window, illustrating the regime where the low-order kinematic assumption fails (Appendix~\ref{app:results-stress}). Higher-order motion models or learned post-collision dynamics are natural extensions. RAFT-derived flow captures only image-plane motion, so scenes with significant depth motion would benefit from depth-augmented flow or multi-view inputs. The physical evaluation uses $\approx 200$ in-lab xArm~7 trajectories, and transfer to substantially different platforms (bimanual systems, humanoids) is untested. The $S$ trajectory samples capture dynamics-model uncertainty but share a single RAFT velocity estimate; sample variance can therefore be confidently wrong when flow is systematically wrong, and folding RAFT confidence into the velocity noise scale, together with epistemic ensembles, would improve calibration. Specular and textureless surfaces, where flow is unreliable, and wrist-mounted cameras, whose ego-flow would need to be removed via forward kinematics, are untested. Predicted targets never bypass the workspace and joint-limit guards, but arm speed is not yet gated by prediction variance. \methodAbbr{} also treats the underlying VLA as fixed; a faster VLA backbone would directly benefit the predict-then-act wrapper, and we leave the joint optimization of backbone and world model to future work.

%% file: appendix/appendix-A0-Architecture-Pipeline.tex
\section{Full Architecture Pipeline}
\label{app:method-architecture-figure}
\input{figures/architecture_full}

%% file: figures/architecture_full.tex
\begin{figure}[!htbp]
    \centering
    \includegraphics[width=0.88\linewidth]{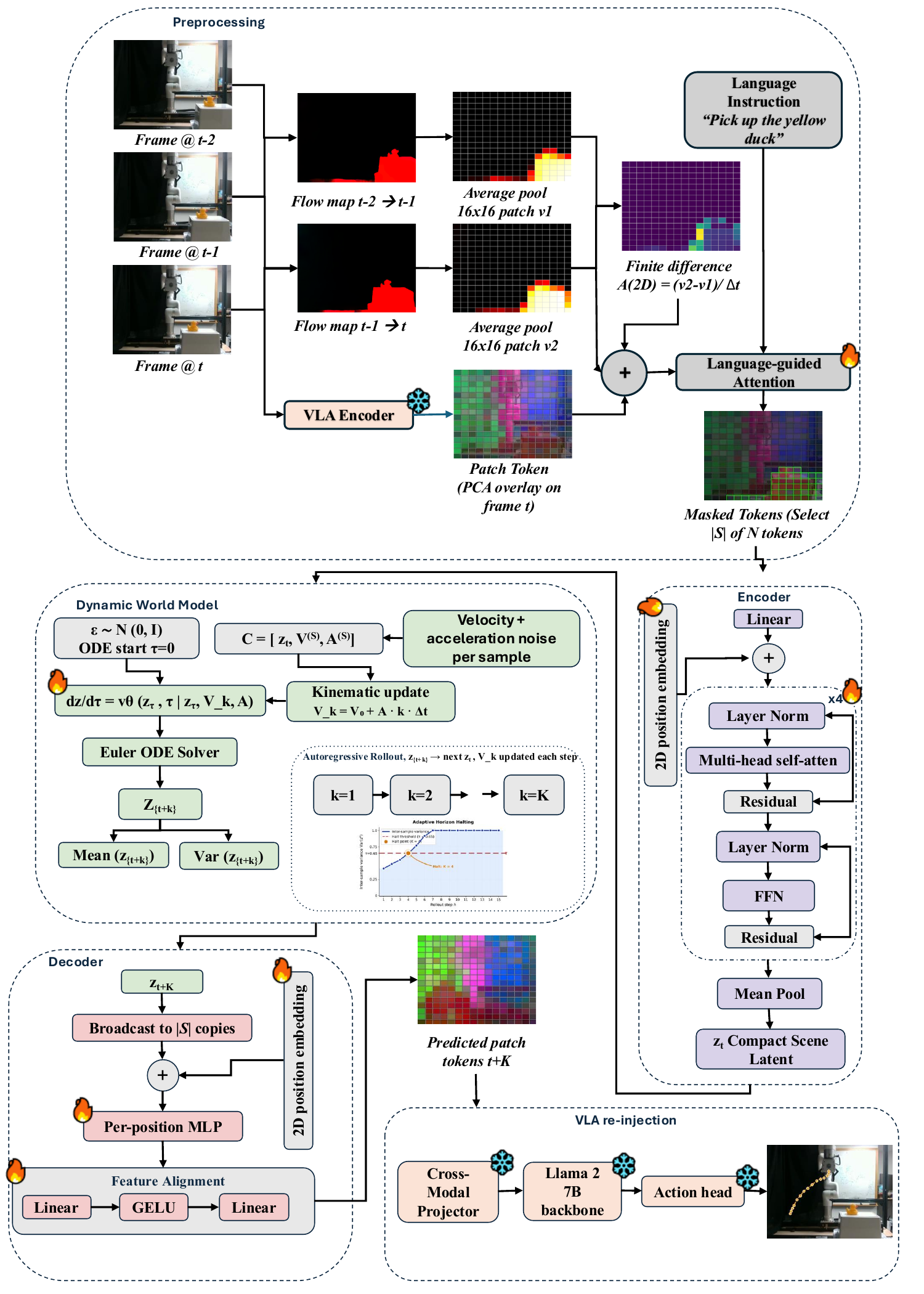}
    \caption{
    \textbf{\methodAbbr{} full architecture.}
    The preprocessing stage computes optical flow between three consecutive frames using RAFT~\citep{Teed2020RAFT}, pools to per-patch velocity, and finite-differences to recover per-patch acceleration. A frozen OpenVLA vision encoder produces $N$ patch tokens, which are gated by language-guided cross-attention to select a task-relevant subset $\mathcal{S}$ of $|\mathcal{S}|$ tokens (typically $30$ to $60$ for $N{=}196$).
    A 4-layer transformer encoder compresses the selected motion-enriched tokens into a compact scene latent $z_t$. The dynamic world model rolls forward via conditional flow matching with analytical kinematic conditioning $V_k = V_0 + A \cdot k \cdot \Delta t$, drawing $S{=}5$ samples per step. The rollout halts when inter-sample variance exceeds threshold $\tau_u$, or at $K_{\max}$ steps if the threshold is never crossed. The decoder reconstructs predicted patch tokens $\hat{v}_{t+K}^{(\mathcal{S})}$, which are spliced into the full $N$-token grid (unselected positions pass through unchanged) and injected into the frozen OpenVLA backbone to produce the action $a_t = \pi(\hat{v}_{t+K}, e_\ell)$.
    Components trained from scratch ($\sim$$4.9$M parameters total) are marked with a flame icon; components inherited frozen from OpenVLA are marked with a snowflake icon.
    }
    \label{fig:architecture-full}
\end{figure}

%% file: appendix/appendix-A-method-details.tex
\section{Method Details}
\label{app:method}

This appendix expands on Section~\ref{sec:method}. Section~\ref{app:method-architecture} gives encoder, decoder, and dynamics-model dimensions. Section~\ref{app:method-alternatives} compares conditional flow matching against diffusion and dropout-based GRU dynamics. Section~\ref{app:method-latency} reports the per-component latency breakdown. Section~\ref{app:method-spatial} details the language-and-motion saliency module. Section~\ref{app:method-motion} details RAFT-based motion estimation. Section~\ref{app:method-kinematic} derives the constant-acceleration kinematic update. Section~\ref{app:method-alignment} describes feature-alignment-layer training. Section~\ref{app:method-training} gives the full training protocol.
\subsection{Architecture Details}
\label{app:method-architecture}

\methodAbbr{} adds $\approx 4.9$M trainable parameters on top of the $\approx 7$B frozen OpenVLA~\citep{Kim2024OpenVLA} backbone, distributed across components as follows: world model encoder $\approx 2.6$M, flow matching dynamics $\approx 1.1$M, decoder $\approx 0.8$M, language-and-motion saliency module $\approx 0.3$M, and per-token feature alignment layer $\approx 0.1$M. The auxiliary modules thus constitute under $0.1$\% of the total parameter count, so the predict-then-act loop is a lightweight wrapper around an unmodified OpenVLA.

\paragraph{Encoder.}
A 4-layer transformer with 256-dim hidden state compresses the selected motion-enriched tokens into a compact latent. Each selected token receives a learnable 2D positional embedding indexed by its $(\text{row}, \text{col})$ patch position, so spatial positions are consistent with the frozen VLA's representation,
\begin{equation}
\label{eq:encoder}
    z_t = g_{\text{enc}}\!\bigl(\tilde{v}_t^{(\mathcal{S})}\bigr)
    \in \mathbb{R}^{d_z},
\end{equation}
where $\tilde{v}_t^{(\mathcal{S})}$ denotes the motion-enriched tokens indexed by $\mathcal{S}$, and $d_z$ is the latent dimension.

\paragraph{Decoder.}
An MLP decoder reconstructs VLA-compatible patch tokens from the predicted latent,
\begin{equation}
\label{eq:decoder}
    \hat{v}_{t+k}^{(\mathcal{S})} = g_{\text{dec}}(z_{t+k})
    \in \mathbb{R}^{|\mathcal{S}| \times d}.
\end{equation}
The decoder splices aligned predicted tokens into the full $N$-token grid, keeping unmasked tokens from the current observation,
\begin{equation}
\label{eq:grid-reassembly}
    \hat{v}_{t+K}[i] =
    \begin{cases}
        g_{\text{align}}\!\bigl(\hat{v}_{t+K}^{(\mathcal{S})}[i]\bigr)
            & \text{if } i \in \mathcal{S}, \\
        v_t[i]
            & \text{otherwise.}
    \end{cases}
\end{equation}
Static background has not changed between $t$ and $t{+}K$, so the current observation is the correct prediction for those tokens.

\paragraph{What the spliced grid represents.}
The splice is a functional surrogate for the future observation, not a visual reconstruction. Predicted tokens are never relocated across the grid. A ballistic ball traverses $8$ to $15$ patches over the rollout, and the selected tokens encode its future kinematic state at their original positions. The alignment layer (Section~\ref{app:method-alignment}) is trained so that the spliced grid drives the frozen decoder to the action it would produce on the true future frame. Table~\ref{tab:abl-splice} shows that the alignment layer recovers fidelity rather than supplying the policy.

\subsection{Alternative Architectures Considered}
\label{app:method-alternatives}

The world model must produce high-quality predictions and complete within the per-step latency budget. We considered three candidate architectures.

\textbf{Diffusion-based world models}~\citep{Ho2020DDPM} yield high-quality samples but require 100 or more denoising steps, placing each rollout at $\approx 380$~ms, outside the budget.

\textbf{Deterministic GRUs}~\citep{Cho2014GRU} run in $\approx 45$~ms but produce a single-point prediction with no uncertainty estimate. Obtaining one via Monte Carlo Dropout~\citep{Gal2016MCDropout} requires 10 forward passes at $10 \times 45 = 450$~ms, also outside the budget.

\textbf{Conditional flow matching} learns a continuous normalizing flow from noise to the predicted future state distribution, requiring 5 integration steps for high-quality samples ($\approx 50$~ms for $S{=}5$ samples). Multiple samples yield both a mean prediction and a natural uncertainty estimate from trajectory sample variance, at no additional overhead.

\methodAbbr{} adopts conditional flow matching as the dynamics model on this basis.

\subsection{Latency Breakdown}
\label{app:method-latency}

Table~\ref{tab:latency} reports the per-component latency breakdown on H100. The $\Delta t_{\max} \approx 200$~ms budget is consistent with human visuomotor reaction loops~\citep{Wolpert1998MOSAIC} and the operating regime of prior reactive manipulation policies~\citep{Morrison2018Closing, Zhao2023ACT}.

\input{tables/latency-breakdown}

The $\approx 158$~ms figure is the replan interval, not the control rate. Action chunks execute at $30$~Hz, the wrapper adds $\approx 68$~ms over the $\approx 90$~ms reactive loop, and the $3$ to $5$ step lookahead spans that delay. Physical experiments run on a single NVIDIA RTX~4090 at 171~ms per action step.

\subsection{Language-and-Motion Saliency Details}
\label{app:method-spatial}

The frozen VLA's language encoder produces embeddings $e_\ell \in \mathbb{R}^{L \times d_l}$ for instruction $\ell$. A learned cross-attention layer attends from the $N$ token positions of $\tilde{v}_t$ to $e_\ell$, producing per-token relevance scores that a sigmoid activation normalizes to $[0,1]$,
\begin{equation}
\label{eq:spatial-attention}
    \widetilde{M} = \sigma\!\bigl(\mathrm{CrossAttn}(\tilde{v}_t,\; e_\ell)\bigr)
    \in [0,1]^{N}.
\end{equation}
A 1-token morphological dilation expands $\widetilde{M}$ to capture object boundaries.

The motion-saliency floor uses $\alpha_{\text{motion}} = 0.5$ in Equation~\eqref{eq:motion-saliency}. The max operator establishes this floor for motion-salient tokens, so that fast-moving objects pass through even when the instruction does not ground there; tokens with high language relevance are unaffected, since their $\widetilde{M}_i$ already dominates. We calibrate $\tau_{\text{flow}}$ on a held-out set of static lab scenes by measuring the 99th-percentile of per-token velocity magnitude under camera noise alone.

Section~\ref{sec:experiments} ablates the language and motion-saliency signals against uniform masking to characterize the compute-accuracy tradeoff. The saliency layer needs no mask labels and trains through the world-model objective.

\paragraph{Robot-arm patches.}
The motion floor in Equation~\eqref{eq:motion-saliency} also fires on the robot's own moving arm. Arm motion is robot-induced rather than independent, and the world model has no action conditioning with which to predict it. We identify arm patches by projecting the robot's link geometry into the image through forward kinematics and the calibrated camera extrinsics, and mark every patch the projection overlaps. Excluding these patches from $\mathcal{S}$ raises Ballistic Catch from $87.4\%$ to 92.2\% (Table~\ref{tab:abl-splice}). All other results in this paper keep arm patches in $\mathcal{S}$, and we recommend the exclusion as the default configuration going forward.

\subsection{Motion Estimation Details}
\label{app:method-motion}

To condition on per-token kinematic state rather than asking the world model to infer motion from visual features alone, \methodAbbr{} estimates both velocity and acceleration from three consecutive observations using RAFT~\citep{Teed2020RAFT} and finite differencing.

\paragraph{Velocity.}
RAFT computes a dense optical flow field $F_{t-1{:}t} \in \mathbb{R}^{H_o \times W_o \times 2}$ between $o_{t-1}$ and $o_t$. Patch-level average-pooling over each $16 \times 16$ region yields a per-token velocity,
\begin{equation}
\label{eq:velocity-pool}
    V_i = \mathrm{AvgPool}_i\!\bigl(F_{t-1{:}t}\bigr)
    \in \mathbb{R}^{2}, \quad i = 1, \ldots, N.
\end{equation}
Average-pooling suffices because task-relevant objects span multiple patches. Interior patches receive clean velocity signals, and the 2 to 4 boundary patches with mixed signals contribute negligibly to prediction error.

\paragraph{Acceleration.}
A second flow field $F_{t-2{:}t-1}$ between $o_{t-2}$ and $o_{t-1}$ is pooled identically to produce a previous velocity field $V^{\text{prev}}$. Finite-differencing the two velocity fields yields per-token acceleration,
\begin{equation}
\label{eq:acceleration}
    A_i = \frac{V_i - V_i^{\text{prev}}}{\Delta t}
    \in \mathbb{R}^{2}, \quad i = 1, \ldots, N,
\end{equation}
where $\Delta t$ is the inter-frame interval. The pipeline concatenates each token's velocity and acceleration into a motion descriptor $[V_i;\, A_i] \in \mathbb{R}^4$ and appends it to the patch token to form the motion-enriched representation $\tilde{v}_t \in \mathbb{R}^{N \times (d+4)}$.

\paragraph{Why per-token and why both signals.}
Per-token motion descriptors preserve spatial structure, so the world model can reason that a specific token contains a moving object with specific velocity and acceleration, where a global motion summary would lose the spatial binding between motion and object identity. Velocity alone suffices for constant-motion scenarios such as conveyors and linear pushes, but acceleration is essential for trajectories that speed up or slow down, including objects accelerating under gravity, decelerating against friction, or undergoing post-contact changes. Section~\ref{sec:experiments} ablates velocity-only, acceleration-only, and combined motion conditioning.

\subsection{Kinematic Update Derivation}
\label{app:method-kinematic}

At step $k{=}1$, real RAFT-derived velocity and acceleration condition the model. At steps $k \geq 2$, no new observation is available, so \methodAbbr{} extrapolates the velocity conditioning analytically under constant-acceleration kinematics per Equation~\eqref{eq:kinematic-update}. This propagation captures the dominant trajectory curvature in acceleration-driven scenarios such as gravitational pull and friction decay, without requiring the world model to learn full physics from data.

To capture motion-estimation uncertainty, \methodAbbr{} injects small Gaussian noise $\sigma = 0.05 \cdot \|V_i\|$ into the velocity conditioning independently per sample, causing trajectories to diverge when predictions are sensitive to exact velocity values and to agree when they are robust. The resulting sample variance is predominantly aleatoric spread plus velocity-noise sensitivity, and it correlates with realized prediction error at $r = $~0.78.
The conditioning $V_k$ never depends on model output (Equation~\eqref{eq:kinematic-update}), training rolls out $K_{\text{train}} = 8$ steps under the model's own predictions, and halting cuts the rollout before error compounds.

\subsection{Feature Alignment Layer}
\label{app:method-alignment}

MSE reconstruction treats all feature dimensions equally, but the VLA action decoder is sensitive to some dimensions, such as spatial relationships critical for grasping, and insensitive to others, such as background texture. To correct this systematic distortion, \methodAbbr{} adds a per-token feature alignment layer that maps decoded features to the action decoder's preferred manifold.

We train the alignment layer separately from the dynamics model on paired real and decoded features, directly optimizing action error,
\begin{equation}
\label{eq:alignment-loss}
    \mathcal{L}_{\text{align}} =
    \bigl\| \pi(g_{\text{align}}(\hat{v}_{t+k}), e_\ell)
    - \pi(v_{t+k}, e_\ell) \bigr\|^2.
\end{equation}

For training purposes, we compute the loss using full-grid action predictions: the predicted tokens $g_{\text{align}}(\hat{v}_{t+k}^{(\mathcal{S})})$ are spliced into the current observation grid per Eq.~\eqref{eq:grid-reassembly} before being passed to $\pi$.

The alignment layer is a two-layer MLP with GELU activation and a residual connection, applied per-token. Training uses paired samples $(v_{t+k}, \hat{v}_{t+k})$ drawn from the validation portion of the pretraining corpus, with $v_{t+k}$ encoded from the actual future frame and $\hat{v}_{t+k}$ predicted by the trained world model. The frozen action decoder $\pi$ provides supervision through the action error gradient. Separating alignment training from world-model training avoids the optimization conflict between reconstruction quality (MSE on features) and action accuracy (downstream policy loss).

\subsection{Training Protocol}
\label{app:method-training}

\paragraph{Why manipulation video pretraining.}
Collecting robot trajectories is expensive, which limits the scale of training data for robot-specific world models. The world model operates in VLA feature space, which the frozen encoder $\phi$ has already trained on diverse internet-scale data. Any video containing object motion in manipulation-relevant contexts therefore provides useful training signal, regardless of whether a robot appears in the scene. This motivates a curated pretraining corpus focused on tabletop interactions and hand-object dynamics, rather than broad egocentric video datasets dominated by outdoor activities and whole-body motion that lie far from the deployment distribution.

\paragraph{Phase 1: video encoding (one-time).}
For each sequence in the training corpus, the frozen encoder $\phi$ encodes every frame into patch tokens, and RAFT computes inter-frame flow,
\begin{equation}
\label{eq:encoding}
    v_i = \phi(o_i), \qquad
    V_i = \mathrm{AvgPool}\!\bigl(\mathrm{RAFT}(o_i, o_{i+1})\bigr),
    \quad i = 1, \ldots, N_{\text{frames}}{-}1.
\end{equation}
For egocentric sources (EPIC-Kitchens~\citep{Damen2022EPIC} and Something-Something~V2~\citep{Goyal2017SSv2}), we estimate and subtract the dominant homography from each flow field before pooling, so that residual flow captures independent object motion rather than camera ego-motion. Table~\ref{tab:pretraining-data} summarizes the training data sources.

\input{tables/pretraining-data}

\paragraph{Phase 2: curriculum pretraining.}
Pretraining follows a three-stage curriculum to build motion understanding progressively. Stage 1 uses simple linear motion from Something-Something~V2 and Bridge~V2~\citep{Walke2023BridgeV2}. Stage 2 adds rotational and multi-object motion from EPIC-Kitchens and DROID~\citep{DROID2024}. Stage 3 mixes all sources with domain-similarity weighting. Each sequence receives a sampling weight that is the product of two factors: a feature-distance factor (1 to 5 times based on Fr\'{e}chet distance between the sequence's VLA features and the deployment distributions of Franka in simulation and xArm~7 in the physical lab) and a source-type factor (1 for egocentric sources, 3 for fixed-camera robot and simulation sources). The latter accounts for ego-motion absence and closer match to the deployment setting in fixed-camera sources (CALVIN, DROID, Bridge~V2, and the OXE xArm subset). We uniformly sample $\approx 150$K trajectories from the Open-X Embodiment~\citep{OpenXEmbodiment2024} xArm subset to balance its contribution against the egocentric and other robot sources.

We jointly train the encoder, dynamics model, and decoder. We train the dynamics model with the standard conditional flow matching objective. We train the encoder and decoder with an MSE reconstruction loss over the rollout horizon $K_{\text{train}}$,
\begin{equation}
\label{eq:pretrain-loss}
    \mathcal{L}_{\text{pre}} =
    \frac{1}{K_{\text{train}}}
    \sum_{k=1}^{K_{\text{train}}}
    \bigl\| g_{\text{dec}}(z_{t+k}) - v_{t+k} \bigr\|^2,
    \quad z_{t+k} \sim p\!\left(\cdot \mid z_{t+k-1}, V_k^{(\mathcal{S})}, A^{(\mathcal{S})}\right),
\end{equation}
where $z_{t+k}$ is sampled autoregressively from the dynamics distribution defined in Equation~\eqref{eq:dynamics}, and $V_k$ updates analytically per Equation~\eqref{eq:kinematic-update}. MSE on features suffices during pretraining because if $\hat{v}_{t+k} \approx v_{t+k}$ in feature space, then the action error $\|\pi(\hat{v}_{t+k}, e_\ell) - \pi(v_{t+k}, e_\ell)\|$ is also small. We train the feature alignment layer separately afterward on paired real and decoded features per Equation~\eqref{eq:alignment-loss}.

\paragraph{Sensitivity to corpus size.}
No evaluation scene or object appears in the pretraining corpus. Retraining the world model on a random $25\%$ of the corpus lowers success by 2.9 points averaged over the four constant-velocity scenes.

\paragraph{Phase 3: fine-tuning on robot data.}
A short fine-tuning phase on $\approx 200$ xArm~7 sequences collected in our lab closes the remaining domain gap between OXE xArm trajectories and the specific camera placement, lighting, and gripper of the deployment setup. Fine-tuning uses the same MSE loss for 5 epochs.

\paragraph{Hyperparameters.}
Table~\ref{tab:hyperparams} lists the key training hyperparameters.

\input{tables/hyperparameters}

%% file: tables/latency-breakdown.tex
\begin{table}[htbp]
    \centering
    \caption{End-to-end latency for one action step on an NVIDIA H100 80GB GPU at $224 \times 224$ input resolution in bfloat16 precision. \methodAbbr{} uses OpenVLA~\citep{Kim2024OpenVLA} with OpenVLA-OFT-style parallel decoding and action chunking for the frozen action head. The full pipeline completes in $\approx 158$~ms, within the $\Delta t_{\max} \approx 200$~ms budget required for reactive manipulation. Per-component values include framework overhead from current PyTorch eager-mode execution and the Python rollout loop; this overhead could be further reduced with compilation (e.g.,~\texttt{torch.compile}, CUDA graphs).}
    \label{tab:latency}
    \begin{tabular}{lc}
        \toprule
        \textbf{Component}
            & \textbf{Latency} \\
        \midrule
        RAFT optical flow~\citep{Teed2020RAFT} ($\times 2$)
            & $\approx 20$~ms \\
        Motion pooling and concatenation
            & $\approx 2$~ms \\
        World model ($S = 5$ flow matching samples)
            & $\approx 40$~ms \\
        Decoder and feature alignment
            & $\approx 6$~ms \\
        VLA forward pass (parallel decoding)
            & $\approx 70$~ms \\
        Framework and I/O overhead
            & $\approx 20$~ms \\
        \midrule
        \textbf{Total end-to-end}
            & $\boldsymbol{\approx 158}$~\textbf{ms} \\
        \bottomrule
    \end{tabular}
\end{table}

%% file: tables/pretraining-data.tex
\begin{table}[htbp]
    \centering
    \small
    \caption{Training data sources for world model pretraining and fine-tuning. The deployment setup uses Franka in simulation and a UFactory xArm~7 in the physical lab. Pretraining uses six public manipulation video sources spanning egocentric human activity and robot demonstrations from multiple embodiments.}
    \label{tab:pretraining-data}
    \begin{tabular}{@{}llrl@{}}
        \toprule
        \textbf{Dataset}
            & \textbf{Content}
            & \textbf{Sequences}
            & \textbf{Usage} \\
        \midrule
        EPIC-Kitchens~\citep{Damen2022EPIC}
            & First-person kitchen manipulation
            & $\approx 50$K
            & Pretrain \\
        Something-Something V2~\citep{Goyal2017SSv2}
            & Hand-object interactions (174 actions)
            & $\approx 220$K
            & Pretrain \\
        DROID~\citep{DROID2024}
            & Diverse robot manipulation (86 tasks)
            & $\approx 76$K
            & Pretrain \\
        Bridge~V2~\citep{Walke2023BridgeV2}
            & WidowX tabletop manipulation
            & $\approx 60$K
            & Pretrain \\
        CALVIN~\citep{Mees2022CALVIN}
            & Simulated Franka manipulation
            & $\approx 100$K
            & Pretrain \\
        OXE~\citep{OpenXEmbodiment2024} (xArm subset)
            & xArm 5/6/7 from OXE
            & $\approx 150$K
            & Pretrain \\
        xArm~7 (in-lab)
            & Physical deployment
            & $\approx 200$
            & Real fine-tuning \\
        \midrule
        \textbf{Total}
            &
            & $\approx 656$K
            & \\
        \bottomrule
    \end{tabular}
\end{table}

%% file: tables/hyperparameters.tex
\begin{table}[!htbp]
    \centering
    \small
    \caption{World model training hyperparameters.}
    \label{tab:hyperparams}
    \begin{tabular}{@{}lr@{}}
        \toprule
        \textbf{Hyperparameter} & \textbf{Value} \\
        \midrule
        Latent dimension $d_z$
            & 256 \\
        Encoder layers / hidden dim
            & 4 / 256 \\
        Encoder attention heads
            & 8 \\
        Flow matching ODE steps
            & 5 (Euler) \\
        Trajectory samples $S$
            & 5 \\
        Maximum horizon $K_{\max}$
            & 10 \\
        Velocity noise $\sigma$
            & $0.05 \times \|V_i\|$ \\
        Uncertainty threshold $\tau_u$
            & 90th-percentile (training) \\
        Optimizer
            & AdamW ($\beta_1{=}0.9$, $\beta_2{=}0.999$) \\
        Learning rate (peak)
            & $3 \times 10^{-4}$ \\
        Learning rate schedule
            & Cosine decay, 2K warmup steps \\
        Weight decay
            & $1 \times 10^{-2}$ \\
        Batch size (pretraining)
            & 256 \\
        Batch size (fine-tuning)
            & 64 \\
        Sequence length (frames)
            & 16 \\
        Training horizon $K_{\text{train}}$
            & 8 \\
        Pretraining steps
            & 200K \\
        Fine-tuning epochs
            & 5 \\
        Gradient clipping (L2 norm)
            & 1.0 \\
        Mixed precision
            & bfloat16 \\
        Hardware
            & 8$\times$NVIDIA H100 80GB \\
        \bottomrule
    \end{tabular}
\end{table}

%% file: appendix/appendix-B-experimental-setup.tex
\section{Experimental Setup}
\label{app:setup}

This appendix expands on the experimental setup summarized in Section~\ref{sec:exp-setup}. Section~\ref{app:setup-sim} details the simulation environments. Section~\ref{app:setup-physical} details the physical xArm~7 setup. Section~\ref{app:setup-baselines} gives baseline implementation details. Section~\ref{app:setup-success} specifies per-scenario success criteria. Section~\ref{app:setup-stats} details the statistical protocol. Section~\ref{app:setup-compute} reports compute resources.

\subsection{Simulation Environments}
\label{app:setup-sim}

All simulation experiments use MuJoCo~\citep{Todorov2012MuJoCo} with a Franka Emika Panda~\citep{Haddadin2022Franka} arm, a parallel-jaw gripper, and a fixed third-person RGB camera rendering at $224 \times 224$ resolution. Control runs at $30$~Hz. Scenes are procedurally generated with randomized object positions, motion parameters, lighting direction, and camera-relative pose.

\paragraph{Constant-velocity scenarios.}
Four scenarios test linear independent motion.

\textit{Conveyor + cup.} A conveyor belt transports a red cup across the workspace at a randomized speed between $5$ and $25$~cm/s. The robot must pick the cup before it leaves the workspace. Belt speed, cup spawn position along the belt, and belt height are randomized per episode.

\textit{Beam + cup.} A red cup slides along a horizontal beam at a randomized speed between $5$ and $20$~cm/s. The robot must grasp the cup from above. Beam orientation and starting position are randomized.

\textit{Pole push + cup.} A pole pushes a cup laterally across the workspace at a randomized speed between $8$ and $22$~cm/s. The robot must intercept the cup. Push direction and starting offset are randomized.

\textit{Rolling ball.} A ball rolls across the table at a constant horizontal velocity between $10$ and $30$~cm/s. The robot must stop the ball.

\paragraph{Acceleration / deceleration scenarios.}
The same four scene topologies, replacing constant velocity with acceleration profiles. Conveyor and beam scenarios add either a linearly increasing or decreasing belt speed over the episode. Pole push adds frictional deceleration. Rolling ball adds gravitational acceleration on an inclined surface (slope randomized between $5$ and $15$ degrees).

\paragraph{Complex dynamic scenarios.}
Eight scenarios test capabilities beyond single-object linear motion.

\textit{Air Hockey (AH).} A puck moves on an air-hockey-style frictionless surface at a randomized initial velocity. The robot, holding a paddle, must deflect the puck.

\textit{Ballistic Catch (BC).} A ball is launched on a parabolic trajectory toward the robot's workspace. The robot must catch the ball before it reaches the floor.

\textit{Multi-Object (MO).} Multiple objects (3 to 5) move simultaneously on linear trajectories. The robot must pick the language-specified object.

\textit{Place in Box (PiB).} A conveyor-mounted box passes through the workspace at $10$ to $20$~cm/s. The robot must place a stationary object into the moving box.

\textit{Multi-Balls (MB).} Three balls launched into the robot workspace. The robot must intercept the language-specified ball.

\textit{Occlusion (Occ).} A launched ball passes behind a vertical occluder for a randomized interval ($300$ to $600$~ms) before reappearing. The robot must catch the object after it emerges.

\textit{Mid-flight trajectory change (Mid).} A launched ball on a parabolic trajectory deflects mid-flight from an unexpected contact, abruptly changing direction. The robot must adjust its intercept target.

\textit{Language + Multi-Object (L+MO).} Three to five moving objects and two boxes of similar appearance but different language descriptions (e.g., ``blue box,'' ``brown box''). The robot must pick the instruction-specified object and bin it in the instruction-specified box.

\paragraph{Stress-test scenarios.}
Three scenarios deliberately exceed the low-order kinematic assumption from Section~\ref{sec:problem}; results are in Appendix~\ref{app:results-stress}.

\textit{Multiple deflection.} A ball bounces off two intermediate surfaces before reaching the workspace.

\textit{Occlusion deflection.} A combination of occlusion and a single deflection during the occluded interval.

\textit{Plinko drop.} A ball falls through a peg array, with each peg contact randomizing its trajectory; the robot must catch the ball at the bottom.

\paragraph{Simulation scene visualizations.}
Figure~\ref{fig:sim-scenes-constant} shows the four constant-velocity scenarios. Figure~\ref{fig:sim-scenes-complex} shows the eight complex scenarios. Figure~\ref{fig:sim-scenes-stress} shows the three stress-test scenarios.

\input{figures/sim-scenes-constant}
\input{figures/sim-scenes-complex}
\input{figures/sim-scenes-stress}

\subsection{Physical Robot Setup}
\label{app:setup-physical}

\paragraph{Robot.}
A UFactory xArm~7 (7-DOF) with a parallel-jaw gripper. The robot operates within a safety bounding box of approximately ${1.2 \times 0.8 \times 0.6}$~m centered in front of the base, with joint-limit guards that halt motion if any joint approaches the manufacturer limits or the workspace bound.

\paragraph{Camera.}
A single Intel RealSense D435 mounted on a fixed third-person tripod approximately ${1.2}$~m from the robot base. Only the RGB stream is used; depth is not consumed by \methodAbbr{} or any baseline. Images are downsampled to $224 \times 224$ to match the OpenVLA encoder input resolution.

\paragraph{Gripper attachments.}
Three end-effector configurations are used across tasks: the bare parallel-jaw gripper for the two conveyor pick-and-place tasks and the rolling-ball task, a ping-pong paddle mounted between the gripper fingers for the paddle-hitting task, and a small soft net mounted to the gripper for the projectile-catching task.

\paragraph{Projectile launcher.}
A spring-loaded mechanical projectile launcher placed ${2}$~m from the robot base, with launch angle and velocity manually varied between trials. The launcher fires soft foam balls.

\paragraph{Conveyor.}
\paragraph{Conveyor.}
A custom belt conveyor with speed adjustable from $0$ to $25$~cm/s in $5$~cm/s increments. The $25$~cm/s ceiling is a hardware limit of the physical conveyor; the simulation speed sweep extends to $40$~cm/s.

\paragraph{Physical setup visualization.}
Figure~\ref{fig:physical-tasks} shows representative frames from each of the four physical tasks.

\input{figures/physical-tasks}

\subsection{Baseline Implementation Details}
\label{app:setup-baselines}

\paragraph{Open-loop VLA.}
OpenVLA~\citep{Kim2024OpenVLA} is queried once on the initial observation, and the predicted action sequence is executed open-loop until completion. No further observations are consumed.

\paragraph{Retargeting VLA (closed-loop).}
OpenVLA is re-queried at every $30$~Hz control step with the latest observation. The first action token from each query is executed before the next query begins, in a fully reactive mode without action chunking.

\paragraph{VLA + Fast Replan.}
OpenVLA with action chunking following the OpenVLA-OFT design~\citep{Kim2025OpenVLAOFT}, with chunk size $K{=}8$ and re-planning at the end of each chunk. The chunk size and replanning interval are tuned on a held-out set of training scenarios to maximize success on dynamic tasks.

\paragraph{Realtime ACT.}
The Action Chunking Transformer~\citep{Zhao2023ACT} adapted for real-time execution. The encoder consumes the current image and proprioception; the decoder predicts a chunk of $32$ future actions. Temporal ensembling averages overlapping predictions from successive chunks. Re-planning occurs every ${8}$ control steps.

\paragraph{Streaming Diffusion Policy.}
Diffusion Policy~\citep{Chi2023DiffusionPolicy} with ${10}$ denoising steps and streaming action-chunk re-planning every ${4}$ control steps.

\paragraph{DreamVLA.}
\citet{Zhang2025DreamVLA} introduce a VLA framework that predicts comprehensive world knowledge as intermediate tokens before generating actions. Their model jointly forecasts dynamic regions, depth maps, and semantic features (extracted from DINOv2 and SAM) using a block-wise structured attention mechanism that prevents information leakage between modalities. We use the released DreamVLA implementation and weights (built on OpenVLA-OFT), fine-tuned on our scenario distribution with the same fine-tuning budget as our other baselines.

\paragraph{Track-and-Intercept VLA.}
A classical motion-informed baseline in which a tracker follows the target object's centroid and fits a constant-acceleration model to its recent positions. The centroid is extrapolated over the same horizon \methodAbbr{} realizes, and the current frame is warped so the object appears at its extrapolated location before the frame enters the same frozen OpenVLA. This is a pixel-space analogue of our latent splice. The tracker is CoTracker3, initialized on the largest connected region whose RAFT flow magnitude exceeds $\tau_{\text{flow}}$. The constant-acceleration fit is a least-squares fit over the last $5$ frames. In multi-object scenes the tracker follows that same region regardless of the instruction, so it has no language grounding.

\paragraph{LaDi-WM (OpenVLA feature space).}
We retrain the LaDi-WM~\citep{Huang2025LaDiWM} latent diffusion world model to predict in OpenVLA feature space, replacing its original feature encoders with the frozen OpenVLA vision encoder $\phi$. It predicts all $N{=}196$ patch tokens at a fixed horizon of $K{=}5$ steps with $20$ denoising steps, receives no optical-flow or kinematic conditioning, and feeds the predicted tokens to the same frozen action decoder as \methodAbbr{}. Training uses the same pretraining corpus (Table~\ref{tab:pretraining-data}) and the same fine-tuning data as our world model, for approximately $96$ GPU-hours on the same hardware.

\paragraph{Training data.}
Simulation demonstrations are scripted rollouts from a privileged-state planner, $500$ per scenario. Every baseline is fine-tuned on these demonstrations with tuned hyperparameters and a shared episode schedule per seed (Section~\ref{app:setup-stats}). The $\approx 200$ real trajectories are static teleoperated pick-and-place episodes plus passive conveyor and launcher clips, so grasping is learned on static scenes and prediction alone supplies the dynamics. The world-model pretraining corpus (Table~\ref{tab:pretraining-data}) is passive video. DreamVLA is likewise video-pretrained before a matched fine-tuning budget (Section~\ref{app:setup-compute}).

\paragraph{Runtime.}
The six learned baselines run at 12 to 95~ms per action step, below \methodAbbr{}'s $\approx 158$~ms, so none is compute-disadvantaged and the reactive ones act on fresher frames.

\paragraph{Hyperparameters.}
Each baseline's training and inference hyperparameters are tuned on a held-out set of ${20}$ training scenarios drawn from the same scenario distribution. The reported numbers use the best-performing hyperparameter configuration per baseline. We report the same fine-tuning protocol (5 epochs on the $\approx {200}$ xArm~7 lab sequences) for all baselines on physical tasks.

\subsection{Success Criteria}
\label{app:setup-success}

Each scenario has scenario-specific pass/fail criteria, evaluated automatically in simulation and by hand on the physical robot.

\paragraph{Pick-and-place scenarios (conveyor + cup, beam + cup, pole push + cup, MO, L+MO).}
Success requires the gripper to close on the target object with all fingers in contact, lift the object by at least ${10}$~cm, and hold it for ${3}$ seconds without dropping. Grasping the wrong object in MO or L+MO scenarios counts as a failure even if the grasp itself succeeds.

\paragraph{Rolling ball.}
Success requires the gripper to physically stop the ball, with the ball's speed falling below 10\% of its pre-contact speed within 500~ms of first contact.

\paragraph{Air hockey.}
Success requires the puck to be deflected back after being struck by the paddle.

\paragraph{Ballistic catch and projectile catch.}
Success requires the ball to be inside the net at the moment its vertical velocity first becomes positive (i.e., the ball has been caught before it bounces off the gripper or the floor).

\paragraph{Place in Box (PiB).}
Success requires the object to be inside the box boundary when the box leaves the workspace, with the object not having touched the table at any point.

\paragraph{Occlusion (Occ).}
Success requires the gripper to catch the target object after the object reappears from behind the occluder.

\paragraph{Mid-flight trajectory change (Mid).}
Success requires the ball to be inside the net at the moment its vertical velocity first becomes positive, where the catch must occur after the trajectory deflection. Trials in which the catch completes before the deflection are excluded from the count.

\paragraph{Stress tests.}
Each stress-test scenario reports success as a function of available reaction time (time-to-exit window). Success requires the gripper to be in contact with the ball within the specified window.

\paragraph{Physical tasks.}
The first conveyor task (moving ducky, static box) requires picking the ducky off the moving conveyor and placing it in the static box. The second conveyor task (static ducky, moving box) requires picking the static ducky and placing it in the box as the box moves along the conveyor, with the ducky landing inside as the box moves the intercept location. The rolling-ball task requires the gripper to stop the ball before it falls off the table edge. The paddle-hitting task requires deflecting the projectile so that it travels at least 30~cm away from the robot after contact. Re-scoring all trials under the quantitative rolling-ball and paddle criteria changes no outcome. The projectile-catching task requires the ball to remain inside the net after first contact, without bouncing out.

\subsection{Statistical Protocol}
\label{app:setup-stats}

\paragraph{Simulation.}
Each cell in the simulation results tables reports the mean success rate across 5 independent training seeds, with $100$ evaluation rollouts per seed, for a total of $500$ rollouts per cell. Standard error is computed across the 5 seed means rather than across the $500$ individual rollouts, so the reported $\pm$ values reflect seed-level variability. All baselines and \methodAbbr{} use the same set of 5 seeds.

\paragraph{Physical robot.}
Each physical task reports successful trials out of $30$ attempts. The physical speed-sensitivity sweep reports $10$ attempts per speed point conducted in a single session per method to control for between-session drift.

\paragraph{Trial randomization.}
For each scenario, the per-trial randomization (object positions, motion parameters, lighting in simulation; launcher angle and velocity on hardware) is drawn from a fixed seed schedule shared across all methods, so all methods see the same set of episode instantiations within a seed. This makes the standard errors directly comparable across methods rather than reflecting differences in episode difficulty.

\subsection{Compute Resources}
\label{app:setup-compute}

\paragraph{Training.}
World model pretraining ran on $8 \times$ NVIDIA H100 80GB GPUs for ${200}$K gradient steps, totaling approximately ${96}$ GPU-hours. The feature alignment layer was trained separately on $1 \times$ H100 for approximately ${4}$ GPU-hours. Fine-tuning on the xArm~7 lab sequences ran on $1 \times$ H100 for approximately ${2}$ GPU-hours.

\paragraph{Baseline training.}
Each baseline was trained on the same hardware for a comparable number of total GPU-hours. Fine-tuning DreamVLA on our scenario distribution used approximately ${80}$ GPU-hours, against approximately $102$ GPU-hours in total for \methodAbbr{}; ACT and Streaming Diffusion Policy fine-tuning used approximately ${8}$ GPU-hours per task suite.

%% file: figures/sim-scenes-constant.tex
\begin{figure}[h]
    \centering
    \begin{subfigure}[t]{0.24\linewidth}
        \centering
        \includegraphics[width=\linewidth]{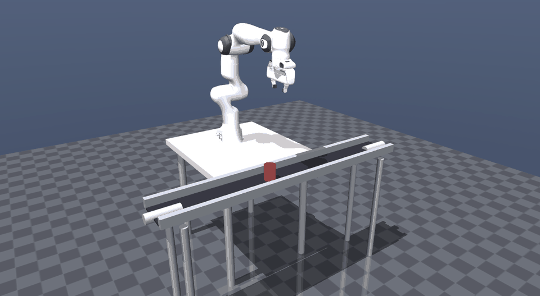}
        \caption{Conveyor + cup}
    \end{subfigure}
    \hfill
    \begin{subfigure}[t]{0.24\linewidth}
        \centering
        \includegraphics[width=\linewidth]{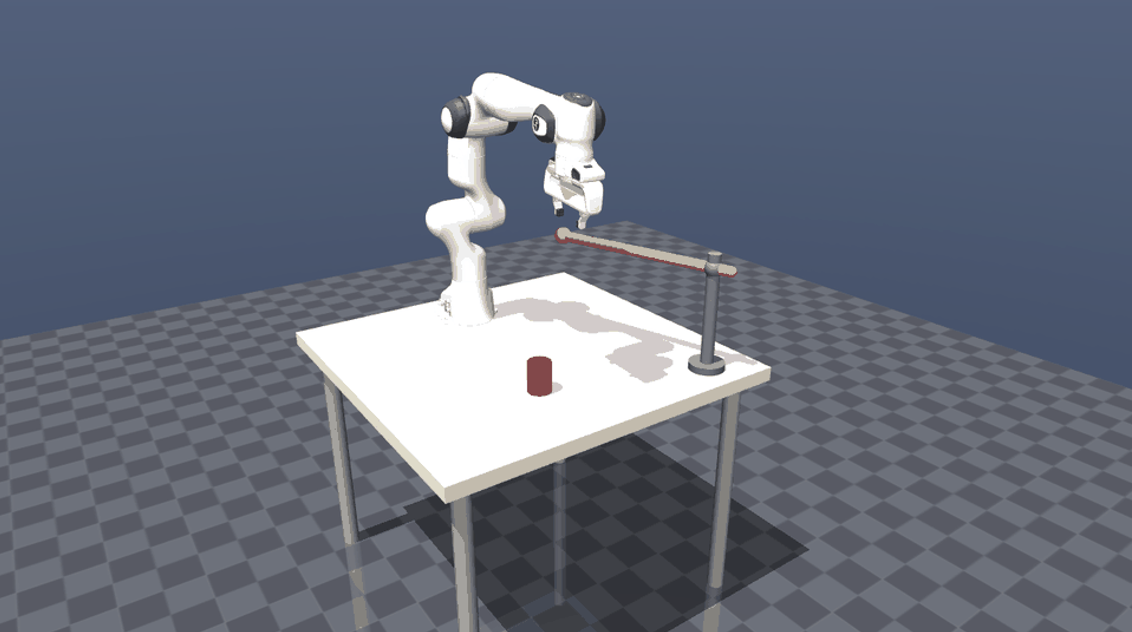}
        \caption{Beam + cup}
    \end{subfigure}
    \hfill
    \begin{subfigure}[t]{0.24\linewidth}
        \centering
        \includegraphics[width=\linewidth]{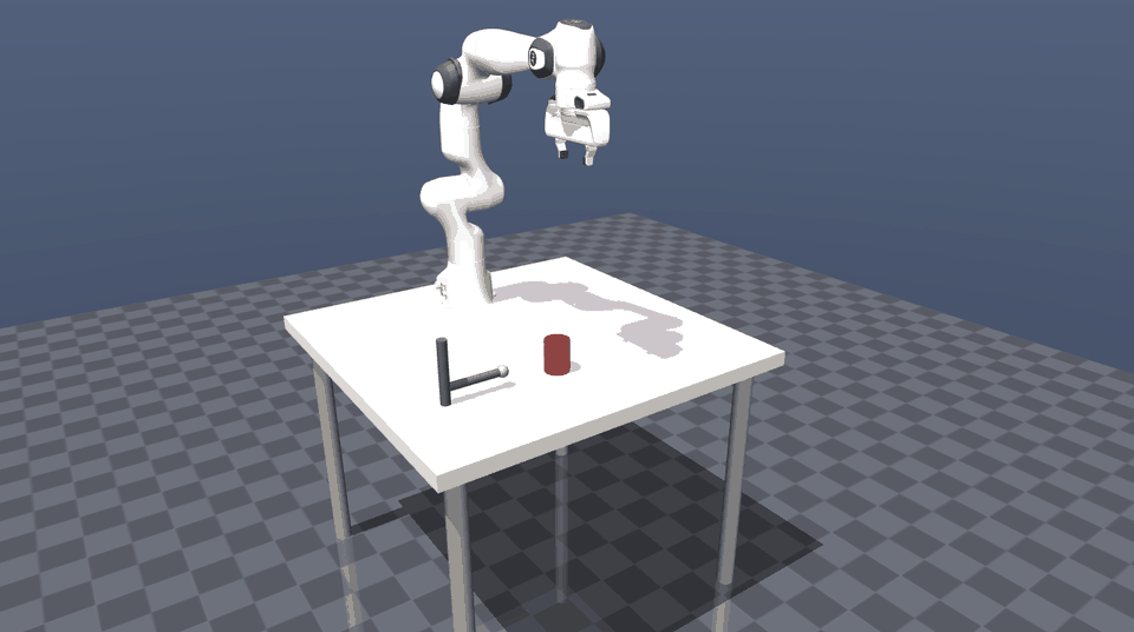}
        \caption{Pole push + cup}
    \end{subfigure}
    \hfill
    \begin{subfigure}[t]{0.24\linewidth}
        \centering
        \includegraphics[width=\linewidth]{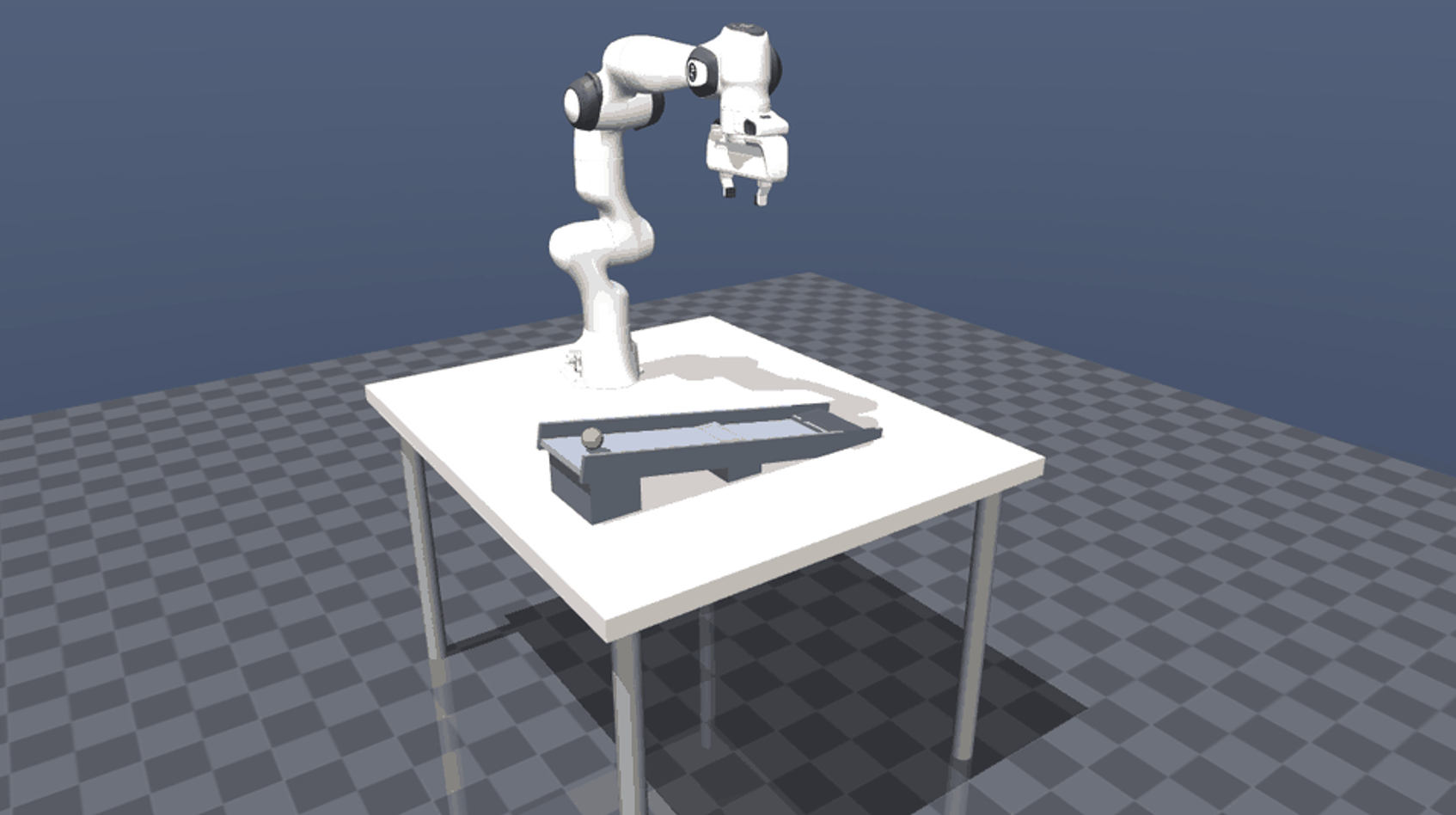}
        \caption{Rolling ball}
    \end{subfigure}
    \caption{Constant-velocity simulation scenarios. The cup or ball moves at a fixed velocity through the workspace; the robot must intercept and grasp.}
    \label{fig:sim-scenes-constant}
\end{figure}

%% file: figures/sim-scenes-complex.tex
\begin{figure}[h]
    \centering
    \begin{subfigure}[t]{0.24\linewidth}
        \centering
        \includegraphics[width=\linewidth]{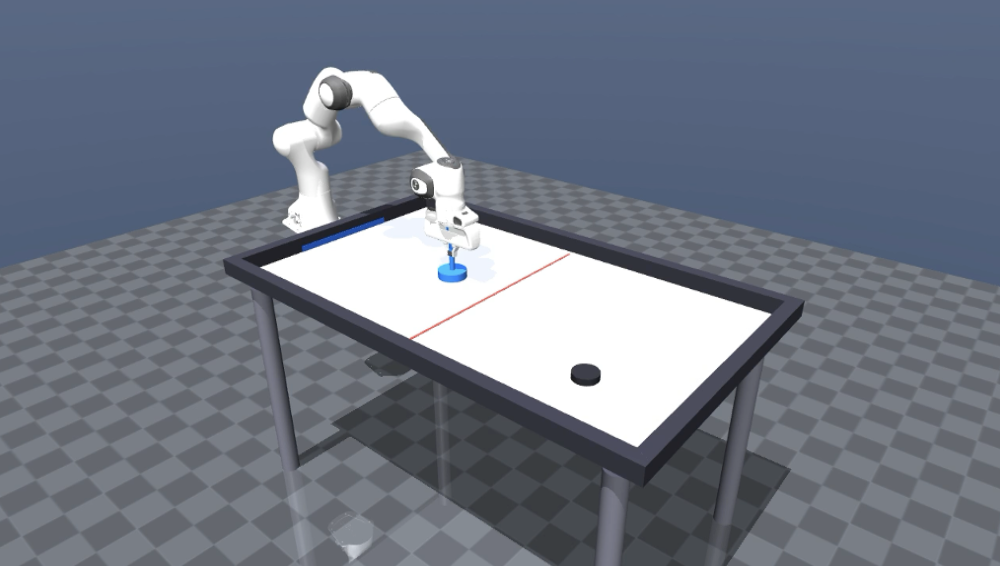}
        \caption{Air Hockey (AH)}
    \end{subfigure}
    \hfill
    \begin{subfigure}[t]{0.24\linewidth}
        \centering
        \includegraphics[width=\linewidth]{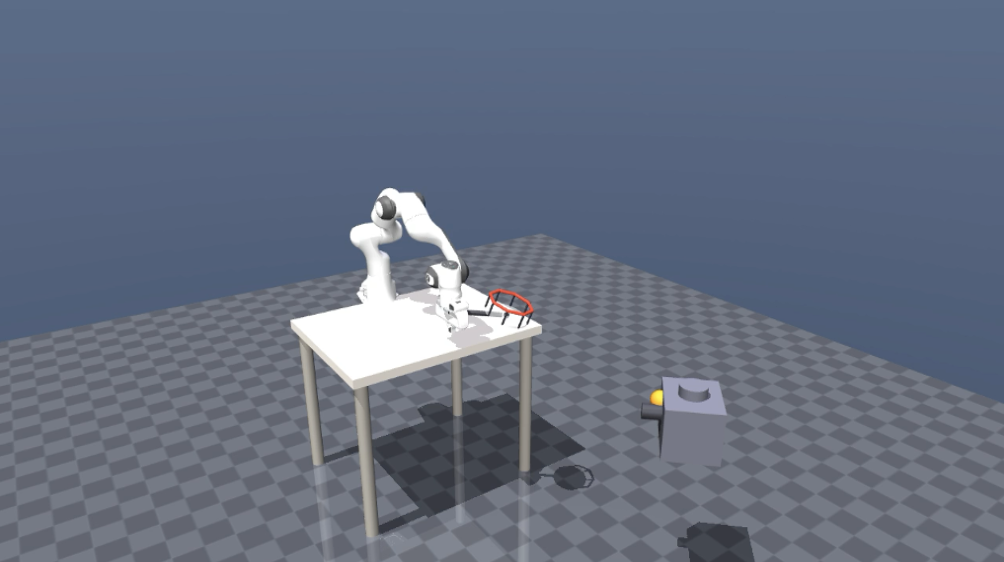}
        \caption{Ballistic Catch (BC)}
    \end{subfigure}
    \hfill
    \begin{subfigure}[t]{0.24\linewidth}
        \centering
        \includegraphics[width=\linewidth]{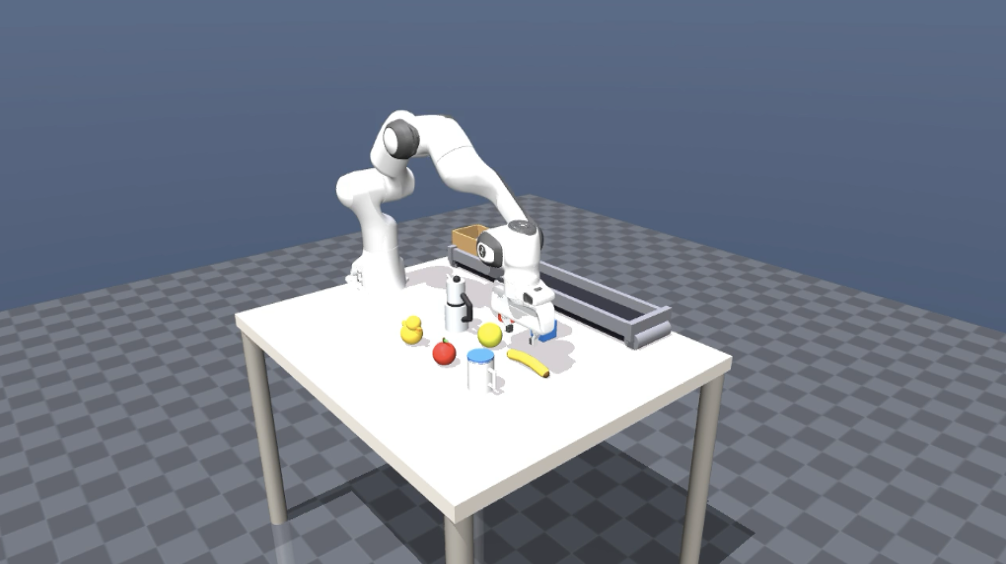}
        \caption{Place in Box (PiB)}
    \end{subfigure}
    \hfill
    \begin{subfigure}[t]{0.24\linewidth}
        \centering
        \includegraphics[width=\linewidth]{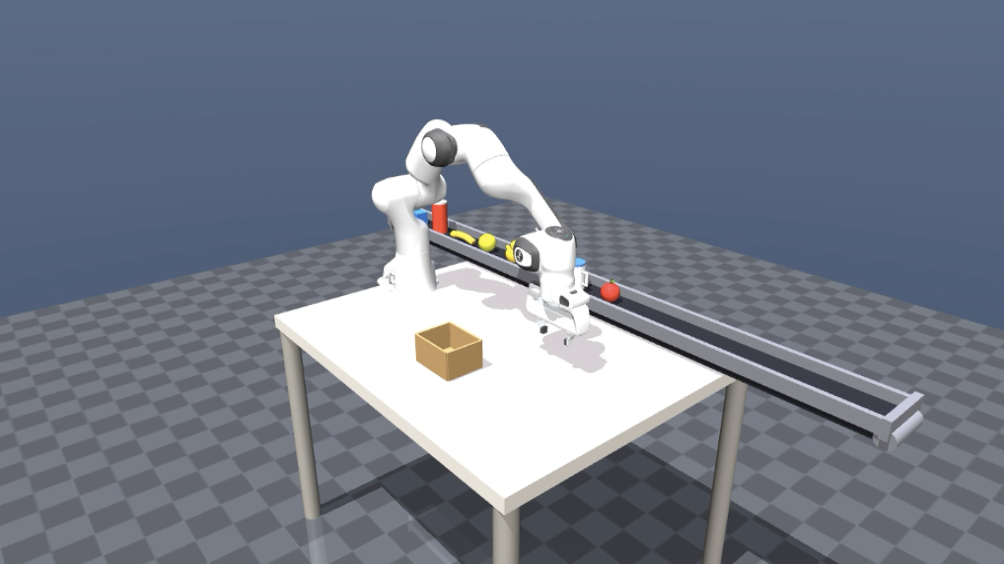}
        \caption{Multi-Object (MO)}
    \end{subfigure}

    \vspace{0.5em}

    \begin{subfigure}[t]{0.24\linewidth}
        \centering
        \includegraphics[width=\linewidth]{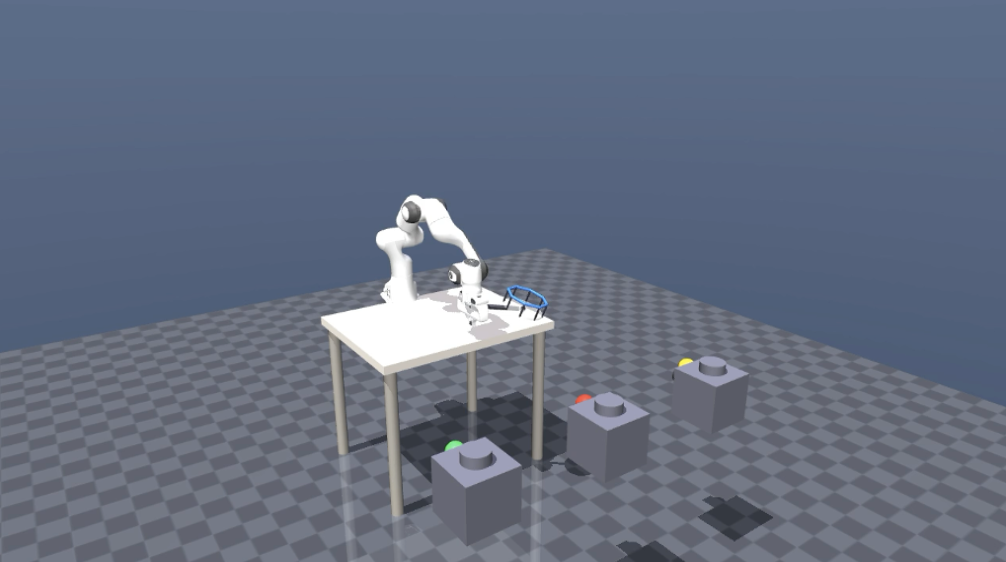}
        \caption{Multi-Balls (MB)}
    \end{subfigure}
    \hfill
    \begin{subfigure}[t]{0.24\linewidth}
        \centering
        \includegraphics[width=\linewidth]{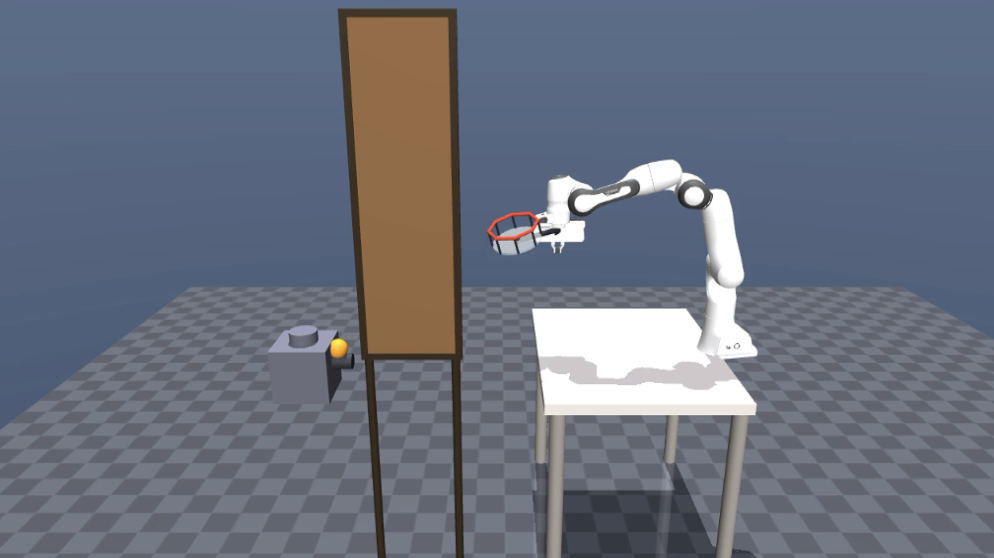}
        \caption{Occlusion (Occ)}
    \end{subfigure}
    \hfill
    \begin{subfigure}[t]{0.24\linewidth}
        \centering
        \includegraphics[width=\linewidth]{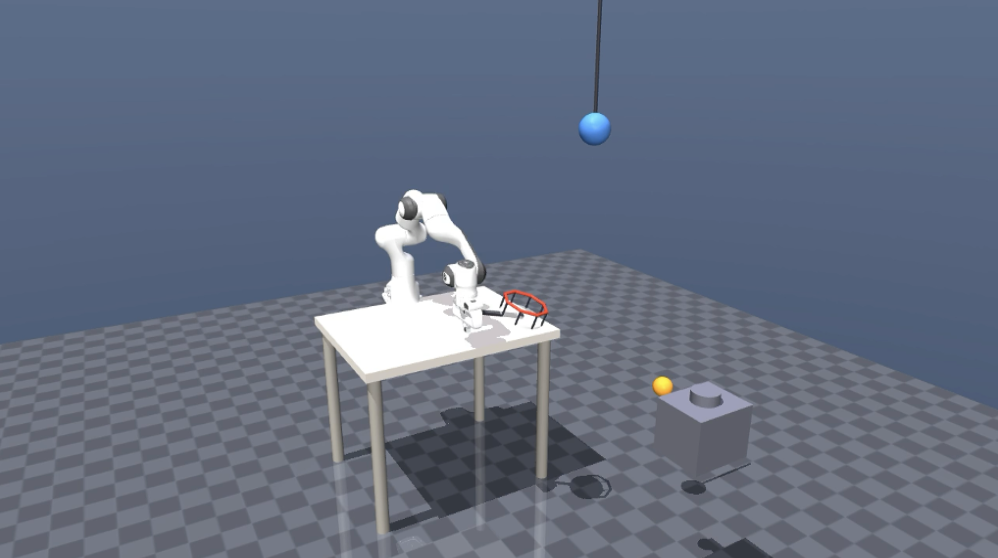}
        \caption{Mid-flight $\Delta$ (Mid)}
    \end{subfigure}
    \hfill
    \begin{subfigure}[t]{0.24\linewidth}
        \centering
        \includegraphics[width=\linewidth]{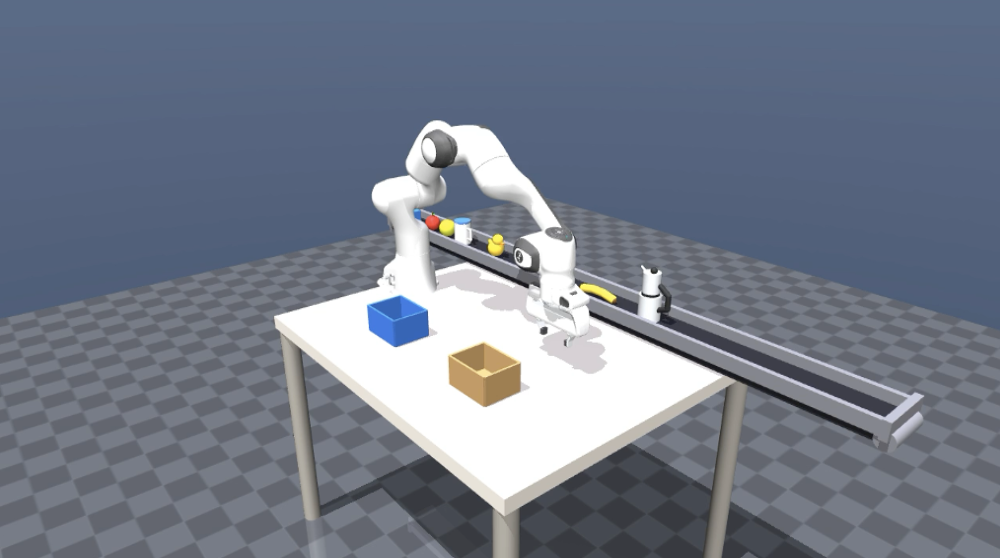}
        \caption{Lang.\ + Multi-obj.\ (L+MO)}
    \end{subfigure}
    \caption{Complex dynamic simulation scenarios. The eight scenes test reactive contact, ballistic trajectories, multi-object selection, occlusion, mid-flight deflection, and language-conditioned target selection.}
    \label{fig:sim-scenes-complex}
\end{figure}

%% file: figures/sim-scenes-stress.tex
\begin{figure}[h]
    \centering
    \begin{subfigure}[t]{0.32\linewidth}
        \centering
        \includegraphics[width=\linewidth]{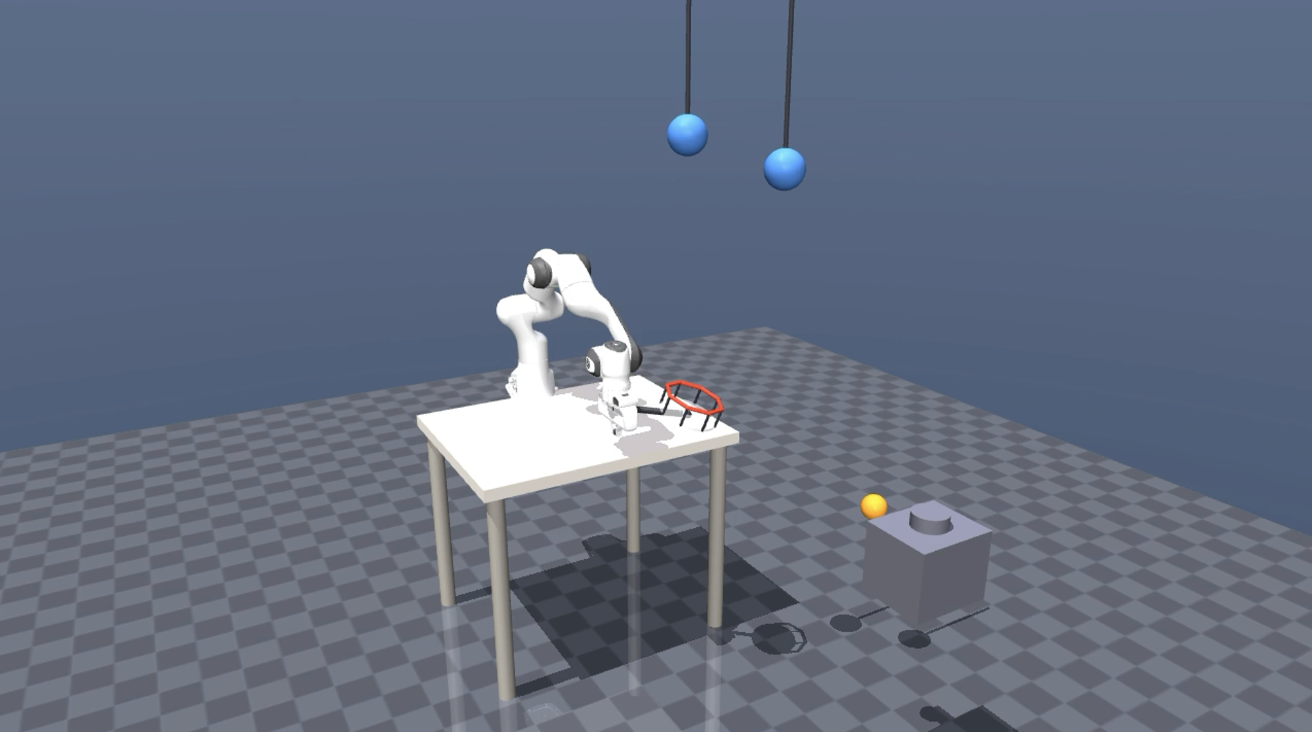}
        \caption{Multiple deflection}
    \end{subfigure}
    \hfill
    \begin{subfigure}[t]{0.32\linewidth}
        \centering
        \includegraphics[width=\linewidth]{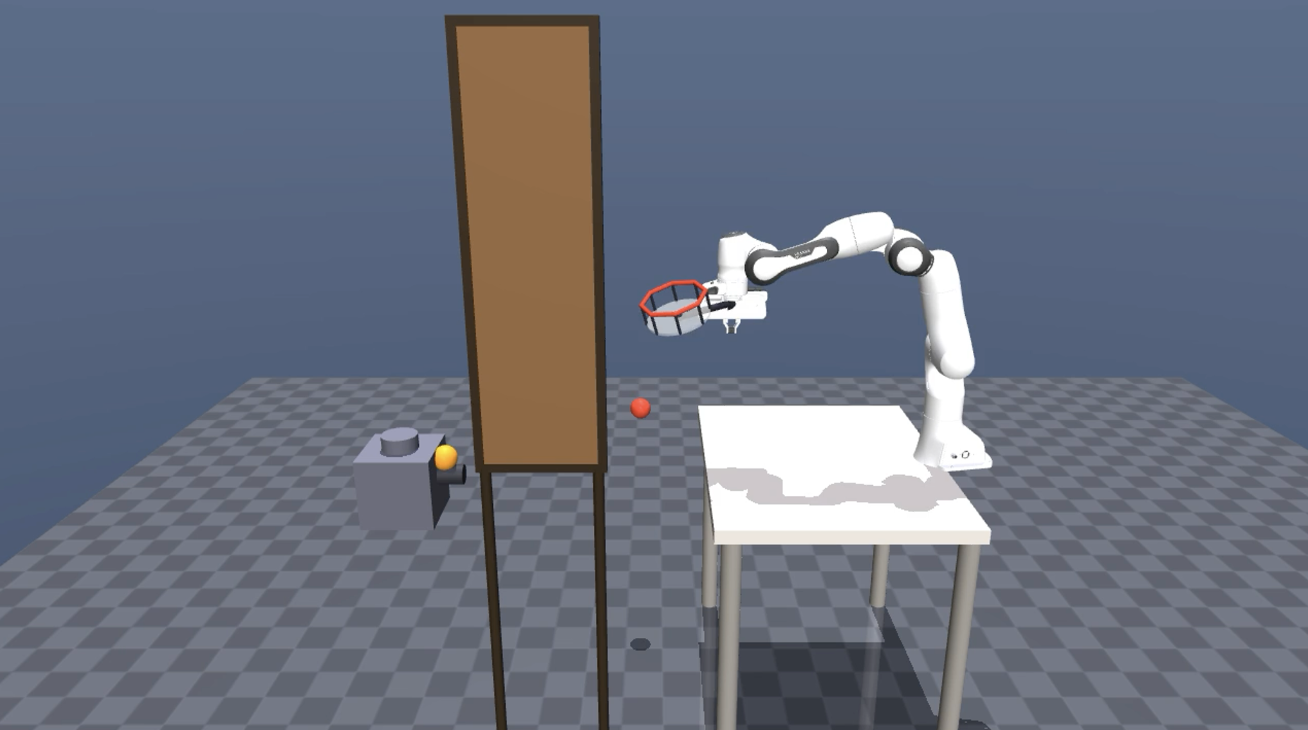}
        \caption{Occlusion deflection}
    \end{subfigure}
    \hfill
    \begin{subfigure}[t]{0.32\linewidth}
        \centering
        \includegraphics[width=\linewidth]{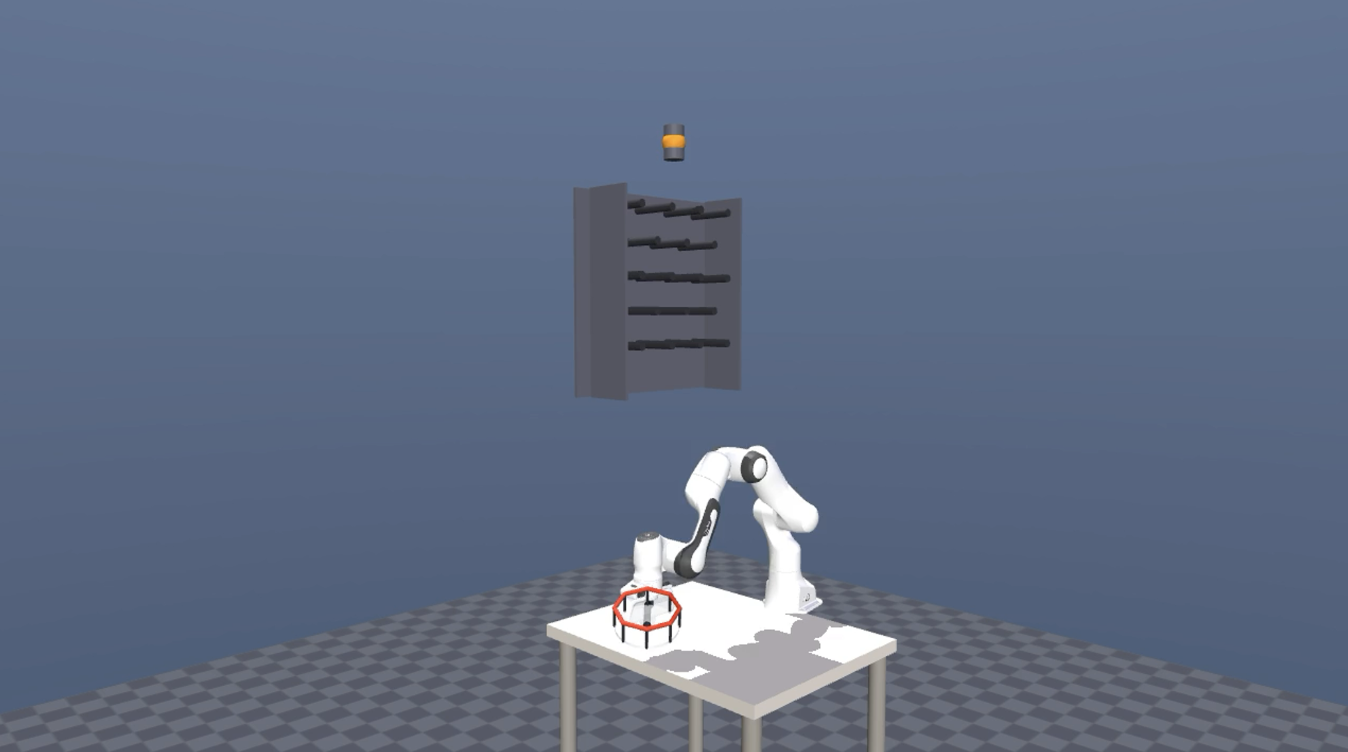}
        \caption{Plinko drop}
    \end{subfigure}
    \caption{Stress-test simulation scenarios that exceed the low-order kinematic motion assumption from Section~\ref{sec:problem}. Each ball undergoes multiple chaotic deflections; \methodAbbr{} degrades gracefully but does not match its in-distribution success rates.}
    \label{fig:sim-scenes-stress}
\end{figure}

%% file: figures/physical-tasks.tex
\begin{figure}[h]
    \centering
    \begin{subfigure}[t]{0.24\linewidth}
        \centering
        \includegraphics[width=\linewidth]{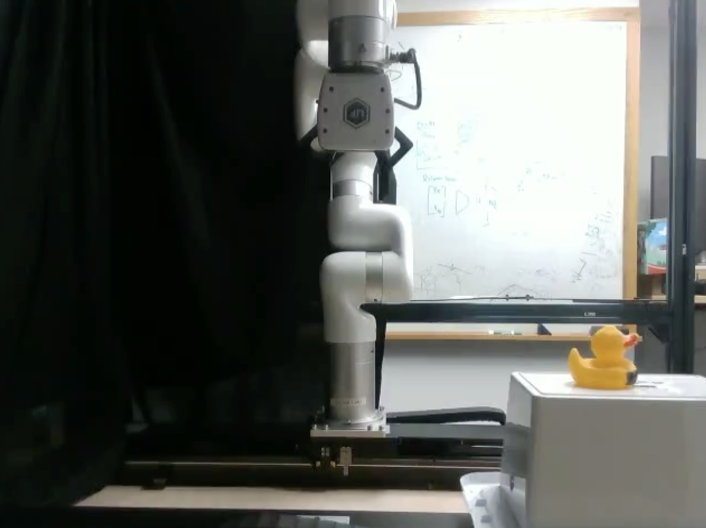}
        \caption{Static box, moving obj.}
    \end{subfigure}
    \hfill
    \begin{subfigure}[t]{0.24\linewidth}
        \centering
        \includegraphics[width=\linewidth]{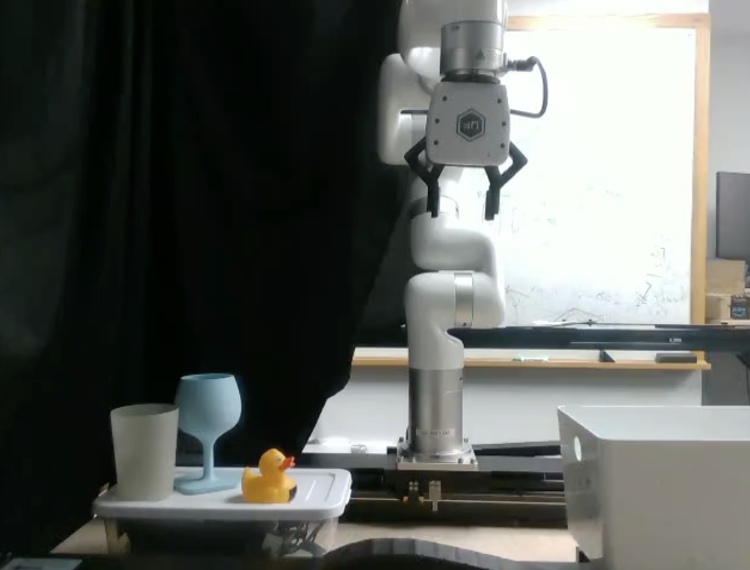}
        \caption{Static obj., moving box}
    \end{subfigure}
    \hfill
    \begin{subfigure}[t]{0.24\linewidth}
        \centering
        \includegraphics[width=\linewidth]{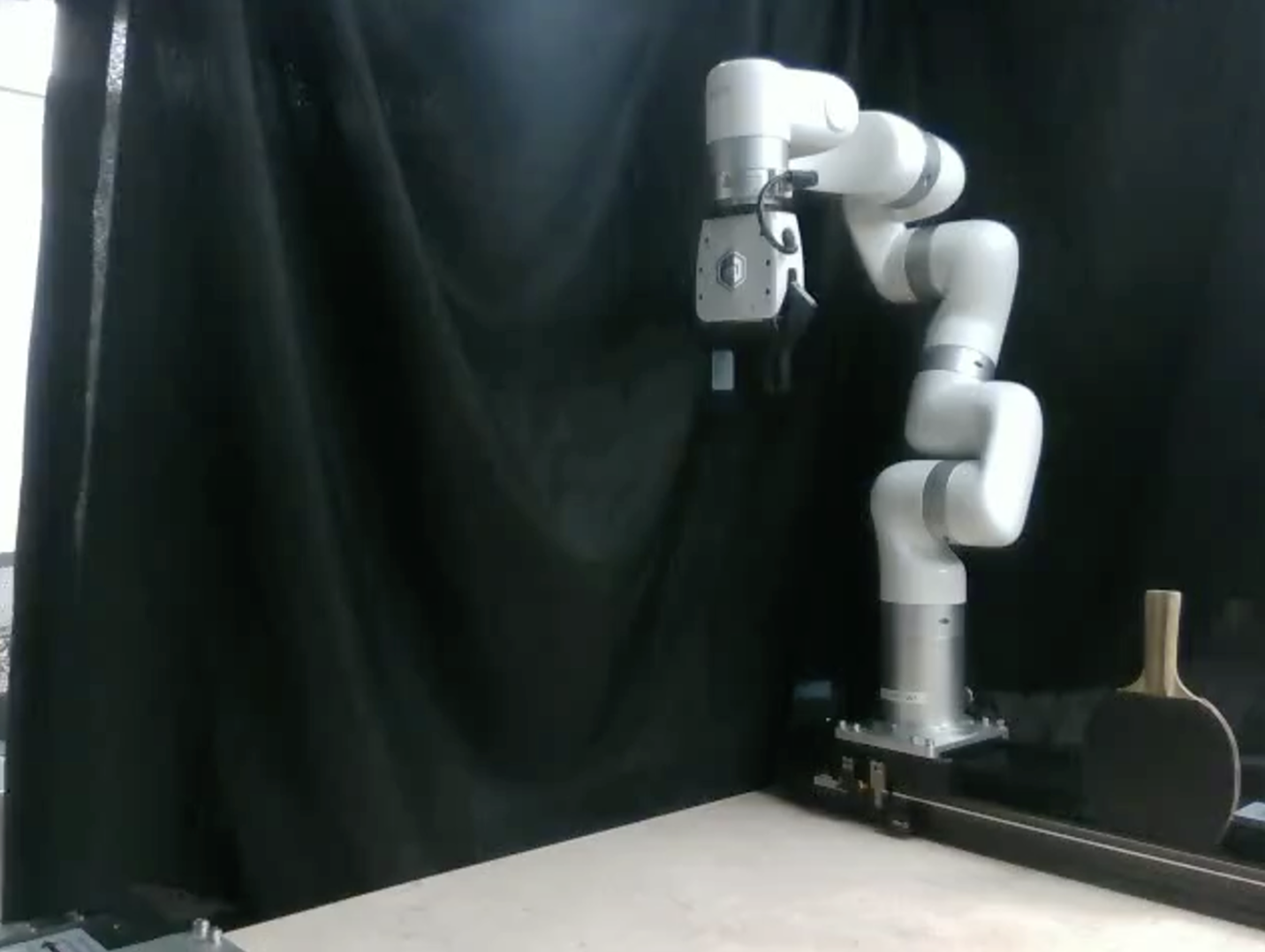}
        \caption{Paddle hitting}
    \end{subfigure}
    \hfill
    \begin{subfigure}[t]{0.24\linewidth}
        \centering
        \includegraphics[width=\linewidth]{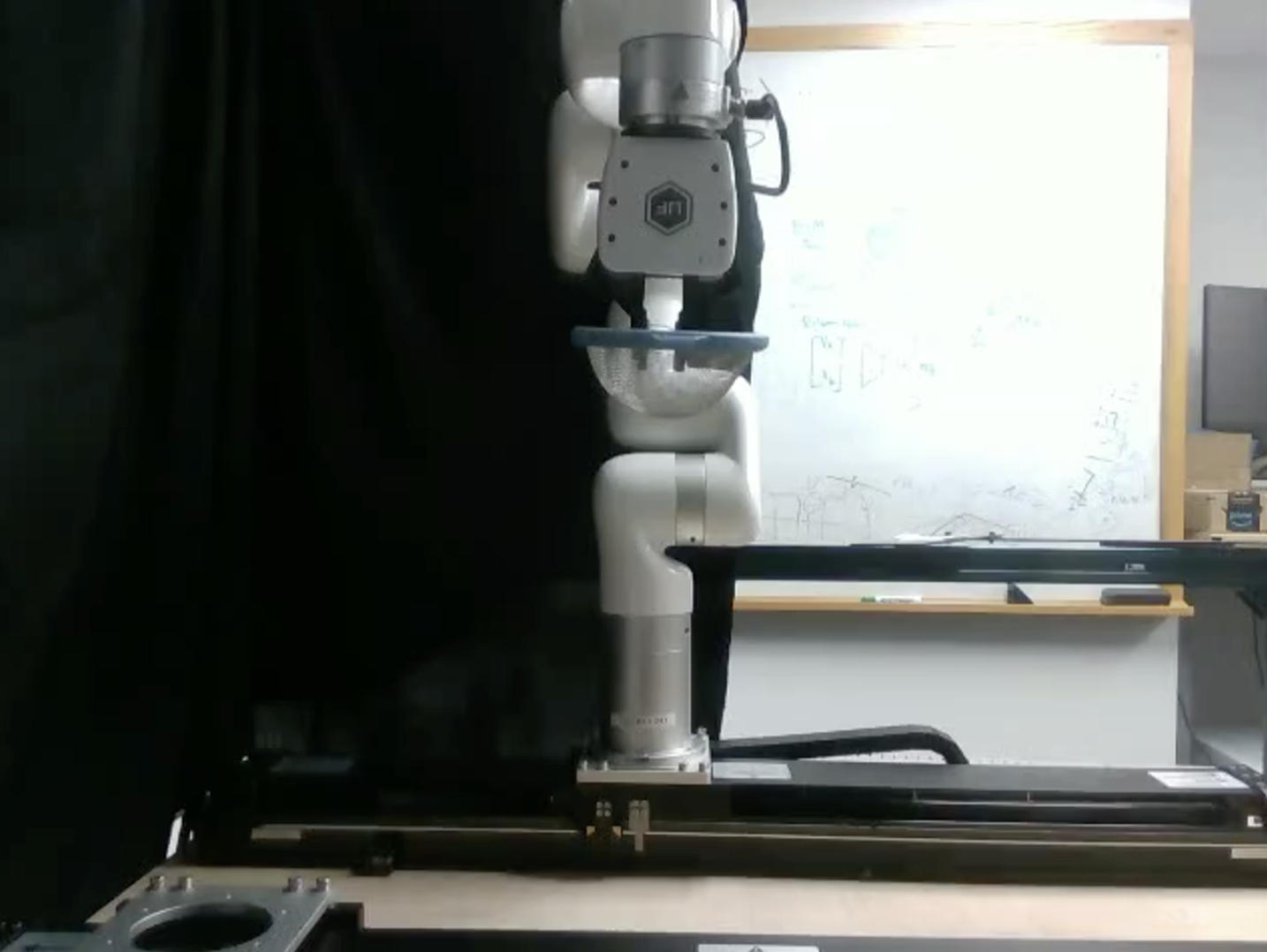}
        \caption{Catch a projectile}
    \end{subfigure}
    \caption{Representative frames from the physical xArm~7 tasks. Left to right: conveyor pick-and-place variants, paddle-based projectile interception, projectile catching with a soft net mounted to the gripper.}
    \label{fig:physical-tasks}
\end{figure}

%% file: appendix/appendix-C-additional-results.tex
\section{Additional Results}
\label{app:results}

This appendix evaluates \methodAbbr{} under stress-test conditions that exceed the operating envelope of the predict-then-act framework.

\subsection{Stress Tests Under Chaotic Dynamics}
\label{app:results-stress}

The Problem Statement's low-order kinematic motion assumption (Section~\ref{sec:problem}) explicitly excludes highly chaotic post-collision dynamics from the operating envelope. This appendix nonetheless evaluates \methodAbbr{} under increasingly chaotic conditions to characterize how the predict-then-act framework degrades when its assumptions are violated. Three stress tests are run, each parameterized by the time-to-exit window during which the robot must intercept the moving object: multiple deflection (a ball bouncing off intermediate surfaces), occlusion deflection (deflection combined with target occlusion), and Plinko drop (a ball falling through a peg array, with the most chaotic dynamics).

\input{tables/results-stress-multi-deflect}

\input{tables/results-stress-occlusion}

\input{tables/results-stress-plinko}

Tables~\ref{tab:stress-multi-deflect} through~\ref{tab:stress-plinko} report task success as a function of available reaction time on the three stress scenarios. At the longest time-to-exit window (1600~ms), \methodAbbr{} achieves $76.4\%$ on multiple deflection, $68.2\%$ on occlusion deflection, and $48.6\%$ on Plinko drop. The Plinko absolute performance is lower than on any other benchmark in this paper, consistent with the fact that Plinko's post-collision dynamics are fundamentally chaotic and violate the low-order kinematic motion assumption. The trajectory after each peg contact is highly sensitive to the contact location and angle, so finite-differenced acceleration provides little predictive signal for steps beyond the next collision. Even in this regime, \methodAbbr{}'s relative advantage over baselines is preserved or amplified. At 1600~ms, \methodAbbr{} achieves $2.7\times$ the success of the strongest baseline on Plinko ($48.6\%$ vs.\ $17.8\%$), $2.1\times$ on multi-deflection, and $2.5\times$ on occlusion deflection. These stress tests bound the operating envelope of predict-then-act control. The claim is graceful degradation under chaotic dynamics, not competitive absolute performance. Adaptive horizon halting contributes to this graceful degradation. The per-scene uncertainty signal truncates rollouts after a peg collision, and Appendix~\ref{app:abl-horizon} compares adaptive against fixed horizon on these scenarios (Table~\ref{tab:abl-horizon-stress}).

%% file: tables/results-stress-multi-deflect.tex
\begin{table}[htbp]
    \centering
    \small
    \caption{Task success rate (\%) on the Multiple Deflection stress test, parameterized by available time-to-exit window. Mean $\pm$ standard error over 5 seeds with 100 rollouts each.}
    \label{tab:stress-multi-deflect}
    \begin{tabular}{@{}lccccc@{}}
        \toprule
        \textbf{Method}
            & \textbf{100 ms}
            & \textbf{200 ms}
            & \textbf{400 ms}
            & \textbf{800 ms}
            & \textbf{1600 ms} \\
        \midrule
        Open-loop VLA              & $0.0 \pm 0.0$ & $0.0 \pm 0.0$ & $0.0 \pm 0.0$ & $0.0 \pm 0.0$  & $0.0 \pm 0.0$ \\
        Retargeting VLA            & $0.0 \pm 0.0$ & $0.0 \pm 0.0$ & $0.0 \pm 0.0$ & $0.6 \pm 1.1$  & $2.4 \pm 2.0$ \\
        VLA + Fast Replan          & $0.0 \pm 0.0$ & $0.0 \pm 0.0$ & $1.2 \pm 1.5$  & $4.8 \pm 2.4$  & $9.2 \pm 2.8$ \\
        Realtime ACT               & $0.0 \pm 0.0$ & $8.6 \pm 2.7$ & $22.4 \pm 4.1$ & $30.4 \pm 4.4$ & $35.6 \pm 4.6$ \\
        Streaming Diffusion Policy & $0.0 \pm 0.0$ & $0.0 \pm 0.0$ & $5.4 \pm 2.4$  & $12.8 \pm 3.2$ & $18.6 \pm 3.7$ \\
        DreamVLA                 & $0.0 \pm 0.0$ & $0.0 \pm 0.0$ & $8.6 \pm 2.8$  & $18.4 \pm 3.8$ & $26.8 \pm 4.2$ \\
        \textbf{\methodAbbr{} (ours)}
                                   & $0.0 \pm 0.0$
                                   & $\mathbf{11.4 \pm 3.0}$
                                   & $\mathbf{35.2 \pm 4.5}$
                                   & $\mathbf{58.6 \pm 4.6}$
                                   & $\mathbf{76.4 \pm 4.1}$ \\
        \bottomrule
    \end{tabular}
\end{table}

%% file: tables/results-stress-occlusion.tex
\begin{table}[htbp]
    \centering
    \small
    \caption{Task success rate (\%) on the Occlusion Deflection stress test, parameterized by available time-to-exit window. Mean $\pm$ standard error over 5 seeds with 100 rollouts each.}
    \label{tab:stress-occlusion}
    \begin{tabular}{@{}lccccc@{}}
        \toprule
        \textbf{Method}
            & \textbf{100 ms}
            & \textbf{200 ms}
            & \textbf{400 ms}
            & \textbf{800 ms}
            & \textbf{1600 ms} \\
        \midrule
        Open-loop VLA              & $0.0 \pm 0.0$ & $0.0 \pm 0.0$ & $0.0 \pm 0.0$ & $0.0 \pm 0.0$  & $0.0 \pm 0.0$ \\
        Retargeting VLA            & $0.0 \pm 0.0$ & $0.0 \pm 0.0$ & $0.0 \pm 0.0$ & $0.4 \pm 0.9$  & $1.8 \pm 1.7$ \\
        VLA + Fast Replan          & $0.0 \pm 0.0$ & $0.0 \pm 0.0$ & $0.6 \pm 1.1$  & $3.4 \pm 2.1$  & $7.2 \pm 2.7$ \\
        Realtime ACT               & $0.0 \pm 0.0$ & $6.4 \pm 2.5$ & $16.2 \pm 3.6$ & $21.6 \pm 4.0$ & $27.4 \pm 4.3$ \\
        Streaming Diffusion Policy & $0.0 \pm 0.0$ & $0.0 \pm 0.0$ & $3.8 \pm 2.9$  & $9.4 \pm 2.8$  & $14.9 \pm 3.4$ \\
        DreamVLA                 & $0.0 \pm 0.0$ & $0.0 \pm 0.0$ & $5.6 \pm 2.5$  & $13.8 \pm 3.4$ & $21.4 \pm 4.0$ \\
        \textbf{\methodAbbr{} (ours)}
                                   & $0.0 \pm 0.0$
                                   & $\mathbf{10.6 \pm 2.9}$
                                   & $\mathbf{30.4 \pm 4.4}$
                                   & $\mathbf{52.8 \pm 4.7}$
                                   & $\mathbf{68.2 \pm 4.4}$ \\
        \bottomrule
    \end{tabular}
\end{table}

%% file: tables/results-stress-plinko.tex
\begin{table}[!htbp]
    \centering
    \small
    \caption{Task success rate (\%) on the Plinko Drop stress test, parameterized by available time-to-exit window. This scenario explicitly violates the constant-acceleration motion assumption (Section~\ref{sec:problem}) and is included to characterize graceful degradation. Mean $\pm$ standard error over 5 seeds with 100 rollouts each.}
    \label{tab:stress-plinko}
    \begin{tabular}{@{}lccccc@{}}
        \toprule
        \textbf{Method}
            & \textbf{100 ms}
            & \textbf{200 ms}
            & \textbf{400 ms}
            & \textbf{800 ms}
            & \textbf{1600 ms} \\
        \midrule
        Open-loop VLA              & $0.0 \pm 0.0$ & $0.0 \pm 0.0$ & $0.0 \pm 0.0$ & $0.0 \pm 0.0$  & $0.0 \pm 0.0$ \\
        Retargeting VLA            & $0.0 \pm 0.0$ & $0.0 \pm 0.0$ & $0.0 \pm 0.0$ & $0.6 \pm 1.1$  & $1.4 \pm 1.5$ \\
        VLA + Fast Replan          & $0.0 \pm 0.0$ & $0.0 \pm 0.0$ & $0.4 \pm 0.9$  & $1.8 \pm 1.7$  & $4.2 \pm 2.3$ \\
        Realtime ACT               & $0.0 \pm 0.0$ & $6.4 \pm 2.5$ & $3.6 \pm 2.9$  & $10.4 \pm 3.3$ & $17.8 \pm 3.8$ \\
        Streaming Diffusion Policy & $0.0 \pm 0.0$ & $0.0 \pm 0.0$ & $0.6 \pm 1.1$  & $3.2 \pm 2.0$  & $8.4 \pm 2.7$ \\
        DreamVLA                 & $0.0 \pm 0.0$ & $0.0 \pm 0.0$ & $1.8 \pm 1.7$  & $6.4 \pm 2.5$  & $12.6 \pm 3.2$ \\
        \textbf{\methodAbbr{} (ours)}
                                   & $0.0 \pm 0.0$
                                   & $\mathbf{9.8 \pm 2.9}$
                                   & $\mathbf{22.6 \pm 4.1}$
                                   & $\mathbf{38.4 \pm 4.6}$
                                   & $\mathbf{48.6 \pm 4.7}$ \\
        \bottomrule
    \end{tabular}
\end{table}

%% file: appendix/appendix-D-ablation-details.tex
\section{Ablation Details}
\label{app:ablations}

This appendix expands the summary in Table~\ref{tab:ablations-summary} with per-scenario results. Each table reports mean $\pm$ standard error over 5 seeds with 100 rollouts each.

\subsection{Motion Estimator}
\label{app:abl-motion}

Table~\ref{tab:abl-motion-estimator} compares RAFT against three alternative motion estimators and a no-velocity baseline on constant-velocity scenes. The implicit-context row feeds the world model the VLA tokens of the same three frames with no flow input and matched capacity. It falls to 58.6\% at $40$~cm/s on conveyor + cup, where \methodAbbr{} holds $95.4\%$. 
Sub-patch displacements are a noisy motion signal in pooled semantic tokens, and without explicit velocity the analytic update of Equation~\eqref{eq:kinematic-update} has nothing to propagate.

\input{tables/ablation-motion-estimator}

\subsection{Velocity Model}
\label{app:abl-velocity}

Table~\ref{tab:abl-velocity-model} ablates the choice of velocity-update model during multi-step rollout on acceleration/deceleration scenes. The constant-velocity assumption performs poorly on conveyor and pole-push scenes ($51\%$ and $58\%$) where objects accelerate or decelerate, but is sufficient for rolling ball under gravity ($91\%$) where dynamics happen to be well-approximated by initial velocity over short rollouts. A fully learned dynamics model yields competitive performance on three scenarios but collapses on rolling ball ($57\%$), where the gravitational acceleration component is most pronounced.

\input{tables/ablation-velocity-model}

\subsection{Spatial Masking}
\label{app:abl-masking}

Table~\ref{tab:abl-masking} ablates the language-and-motion saliency masking strategy on constant-velocity scenes. Full prediction is within error of saliency masking in accuracy, so the mask is accuracy-neutral here. Its contribution is latency. Predicting all $196$ tokens instead of the $30$ to $60$ selected exceeds the $\approx 200$~ms budget, and the mask keeps prediction deployable.

\input{tables/ablation-masking}

\subsection{Fixed Horizon Sweep}
\label{app:abl-horizon}

Table~\ref{tab:abl-horizon} reports a controlled fixed-horizon sweep on constant-velocity scenes, characterizing the latency-accuracy tradeoff that motivates \methodAbbr{}'s adaptive halting mechanism. Outside this subsection, all results use adaptive halting with $K_{\max} = 10$. The optimal fixed horizon varies by scenario. $K = 5$ peaks on conveyor and rolling ball, while $K = 8$ is essentially tied on beam.

\input{tables/ablation-horizon}

Fixed $K{=}5$ ties adaptive halting on these scenes because halting realizes $3$ to $5$ steps there. The benefit of adaptivity appears when predictability varies. Table~\ref{tab:abl-horizon-stress} compares the two on the stress tests of Appendix~\ref{app:results-stress}, where post-collision variance truncates rollouts at the contact, and Table~\ref{tab:realized-horizon} reports the realized horizon per scenario group.

\input{tables/ablation-horizon-stress}

\input{tables/realized-horizon}

\subsection{World Model Architecture}
\label{app:abl-architecture}

Table~\ref{tab:abl-architecture} ablates the number of flow matching samples $S$ and encoder depth on constant-velocity scenes. Reducing $S$ to 1 sample eliminates the trajectory-variance uncertainty signal entirely and drops performance to $58$ to $80\%$. $S = 3$ recovers most of the performance but leaves the uncertainty estimate noisy. $S = 5$ is the best operating point, recovering full performance within the latency budget. $S = 10$ yields marginal further gains at substantially higher compute. A 2-layer encoder is insufficient ($70$ to $78\%$ success), while 6 layers offer essentially no gain over 4 layers at $14$~ms additional cost.

\input{tables/ablation-architecture}

\subsection{Token Splicing}
\label{app:abl-splice}

Table~\ref{tab:abl-splice} ablates two aspects of the token splice in Equation~\eqref{eq:grid-reassembly}. Without the alignment layer, Ballistic Catch remains above the best baseline, so the layer, trained once on the pretraining corpus and never per task, recovers fidelity rather than supplying the policy. Excluding robot-arm patches from $\mathcal{S}$ removes predictions of robot-induced motion that the world model cannot make without action conditioning (Appendix~\ref{app:method-spatial}).

\input{tables/ablation-splice}

%% file: tables/ablation-motion-estimator.tex
\begin{table}[htbp]
    \centering
    \small
    \caption{Motion estimator ablation on constant-velocity scenes. Mean $\pm$ standard error over 5 seeds with 100 rollouts each.}
    \label{tab:abl-motion-estimator}
    \begin{tabular}{@{}lcccc@{}}
        \toprule
        \textbf{Motion estimator}
            & \textbf{Conveyor + cup}
            & \textbf{Beam + cup}
            & \textbf{Pole push + cup}
            & \textbf{Rolling ball} \\
        \midrule
        \textbf{RAFT (ours)}
                          & $\mathbf{97.3 \pm 1.5}$
                          & $\mathbf{91.7 \pm 2.8}$
                          & $\mathbf{95.0 \pm 2.3}$
                          & $\mathbf{90.0 \pm 3.1}$ \\
        CoTracker3       & $89.3 \pm 3.1$ & $87.7 \pm 3.3$ & $80.7 \pm 4.0$ & $78.3 \pm 4.1$ \\
        Implicit 3-frame context & $80.4 \pm 3.9$ & $75.6 \pm 2.2$ & $73.2 \pm 4.4$ & $67.6 \pm 3.6$ \\
        Farneb\"ack       & $79.7 \pm 4.0$ & $69.3 \pm 4.6$ & $68.7 \pm 4.7$ & $58.0 \pm 4.9$ \\
        Frame differencing & $45.0 \pm 5.0$ & $39.7 \pm 4.9$ & $40.3 \pm 2.1$ & $29.0 \pm 4.5$ \\
        No velocity input & $0.0 \pm 0.0$  & $0.0 \pm 0.0$  & $0.0 \pm 0.0$  & $0.0 \pm 0.0$ \\
        \bottomrule
    \end{tabular}
\end{table}

%% file: tables/ablation-velocity-model.tex
\begin{table}[htbp]
    \centering
    \small
    \caption{Velocity model ablation on acceleration/deceleration scenes. Mean $\pm$ standard error over 5 seeds with 100 rollouts each.}
    \label{tab:abl-velocity-model}
    \begin{tabularx}{\linewidth}{@{}X cccc@{}}
        \toprule
        \textbf{Velocity model}
            & \textbf{Conveyor + cup}
            & \textbf{Beam + cup}
            & \textbf{Pole push + cup}
            & \textbf{Rolling (grav.)} \\
        \midrule
        Constant: $V_k = V_0$ & $50.6 \pm 4.3$ & $87.2 \pm 2.8$ & $58.4 \pm 4.7$ & $90.6 \pm 2.4$ \\
        Exponential decay: $V_k = V_0 e^{-\lambda k}$ & $42.8 \pm 5.1$ & $78.4 \pm 3.6$ & $50.2 \pm 4.4$ & $89.8 \pm 2.6$ \\
        Kinematic only (no accel in token) & $78.2 \pm 3.9$ & $86.8 \pm 3.4$ & $74.6 \pm 4.2$ & $90.4 \pm 2.7$ \\
        Accel in token only (no kinematic) & $68.4 \pm 4.5$ & $85.4 \pm 3.5$ & $68.8 \pm 4.6$ & $91.2 \pm 2.5$ \\
        \textbf{Kinematic + accel-conditioned (ours)}
                          & $\mathbf{88.4 \pm 3.1}$
                          & $\mathbf{91.6 \pm 2.6}$
                          & $\mathbf{93.2 \pm 2.3}$
                          & $89.4 \pm 2.9$ \\
        Learned dynamics (accel and velocity) & $76.8 \pm 4.0$ & $89.4 \pm 3.0$ & $91.8 \pm 2.7$ & $56.8 \pm 4.8$ \\
        \bottomrule
    \end{tabularx}
\end{table}

%% file: tables/ablation-masking.tex
\begin{table}[htbp]
    \centering
    \small
    \caption{Spatial masking ablation on constant-velocity scenes, with end-to-end latency per action step. Full prediction matches saliency masking in accuracy but exceeds the $\approx 200$~ms budget. Mean $\pm$ standard error over 5 seeds with 100 rollouts each.}
    \label{tab:abl-masking}
    \setlength{\tabcolsep}{4pt}
    \resizebox{\linewidth}{!}{%
    \begin{tabular}{@{}lccccc@{}}
        \toprule
        \textbf{Masking strategy}
            & \textbf{Conveyor}
            & \textbf{Beam}
            & \textbf{Pole push}
            & \textbf{Rolling ball}
            & \textbf{Latency (ms)} \\
        \midrule
        \textbf{Language + motion (ours)}
                          & $\mathbf{97.3 \pm 1.5}$
                          & $\mathbf{92.4 \pm 2.6}$
                          & $94.6 \pm 2.4$
                          & $\mathbf{91.2 \pm 2.8}$
                          & $158 \pm 2$ \\
        Motion saliency only (no language) & $86.2 \pm 3.4$ & $79.8 \pm 3.9$ & $84.4 \pm 3.5$ & $85.6 \pm 3.4$ & $151 \pm 2$ \\
        Language only (no motion boost) & $87.4 \pm 3.2$ & $68.6 \pm 4.5$ & $79.2 \pm 3.8$ & $76.4 \pm 4.1$ & $154 \pm 2$ \\
        Uniform mask (random $30\%$) & $49.4 \pm 4.6$ & $41.6 \pm 4.8$ & $51.8 \pm 4.5$ & $42.8 \pm 4.7$ & $160 \pm 2$ \\
        Full prediction (all $196$ tokens) & $96.0 \pm 2.2$ & $86.8 \pm 3.3$ & $\mathbf{95.6 \pm 2.4}$ & $86.4 \pm 3.4$ & $250 \pm 3$ \\
        \bottomrule
    \end{tabular}%
    }
\end{table}

%% file: tables/ablation-horizon.tex
\begin{table}[htbp]
    \centering
    \small
    \caption{Fixed-horizon ablation on constant-velocity scenes, characterizing the latency-accuracy tradeoff. Mean $\pm$ standard error over 5 seeds with 100 rollouts each.}
    \label{tab:abl-horizon}
    \begin{tabular}{@{}lccccc@{}}
        \toprule
        \textbf{Horizon $K$}
            & \textbf{Conveyor + cup}
            & \textbf{Beam + cup}
            & \textbf{Pole push + cup}
            & \textbf{Rolling ball}
            & \textbf{Latency (ms)} \\
        \midrule
        $K = 1$  & $51.2 \pm 4.4$ & $59.8 \pm 4.5$ & $48.4 \pm 4.7$ & $40.6 \pm 4.9$ & $120 \pm 1$ \\
        $K = 2$  & $58.6 \pm 4.5$ & $65.4 \pm 4.6$ & $60.2 \pm 4.7$ & $56.8 \pm 4.8$ & $128 \pm 1$ \\
        $K = 3$  & $78.6 \pm 2.9$ & $84.4 \pm 3.6$ & $76.2 \pm 4.1$ & $78.8 \pm 3.9$ & $138 \pm 1$ \\
        $K = 5$  & $\mathbf{96.8 \pm 1.9}$ & $\mathbf{92.4 \pm 2.6}$ & $\mathbf{94.6 \pm 2.4}$ & $\mathbf{91.2 \pm 2.8}$ & $158 \pm 2$ \\
        $K = 8$  & $95.6 \pm 2.2$ & $91.8 \pm 2.8$ & $93.8 \pm 2.6$ & $90.4 \pm 3.0$ & $185 \pm 2$ \\
        $K = 12$ & $88.6 \pm 3.2$ & $91.4 \pm 2.7$ & $87.4 \pm 3.4$ & $77.8 \pm 4.0$ & $218 \pm 3$ \\
        \bottomrule
    \end{tabular}
\end{table}

%% file: tables/ablation-horizon-stress.tex
\begin{table}[htbp]
    \centering
    \small
    \caption{Adaptive halting versus fixed horizon on the three stress tests at the $1600$~ms time-to-exit window. Mean $\pm$ standard error over 5 seeds with 100 rollouts each.}
    \label{tab:abl-horizon-stress}
    \begin{tabular}{@{}lccc@{}}
        \toprule
        \textbf{Horizon}
            & \textbf{Multiple deflection}
            & \textbf{Occlusion deflection}
            & \textbf{Plinko drop} \\
        \midrule
        \textbf{Adaptive halting (ours)} & $76.4 \pm 4.1$ & $68.2 \pm 4.4$ & $48.6 \pm 4.7$ \\
        Fixed $K = 5$                    & $63.8 \pm 3.5$ & $54.6 \pm 6.7$ & $31.4 \pm 1.4$ \\
        \bottomrule
    \end{tabular}
\end{table}

%% file: tables/realized-horizon.tex
\begin{table}[htbp]
    \centering
    \small
    \caption{Mean realized rollout horizon $K$ under adaptive halting ($K_{\max} = 10$), averaged over evaluation rollouts.}
    \label{tab:realized-horizon}
    \begin{tabular}{@{}lccccc@{}}
        \toprule
            & \textbf{Const.\ vel.}
            & \textbf{Accel.\ / decel.}
            & \textbf{Ballistic}
            & \textbf{Occlusion}
            & \textbf{Plinko} \\
        \midrule
        Realized $K$ (steps) & 4.2 & 3.8 & 9.1 & 9.6 & 1.8 \\
        \bottomrule
    \end{tabular}
\end{table}

%% file: tables/ablation-architecture.tex
\begin{table}[htbp]
    \centering
    \small
    \caption{World model architecture ablation on constant-velocity scenes. Mean $\pm$ standard error over 5 seeds with 100 rollouts each.}
    \label{tab:abl-architecture}
    \begin{tabular}{@{}lccccc@{}}
        \toprule
        \textbf{Configuration}
            & \textbf{Conveyor}
            & \textbf{Beam}
            & \textbf{Pole push}
            & \textbf{Rolling ball}
            & \textbf{Latency (ms)} \\
        \midrule
        Flow matching, $S = 1$  & $78.6 \pm 3.8$ & $80.2 \pm 3.7$ & $70.4 \pm 4.4$ & $58.4 \pm 4.6$ & $125 \pm 1$ \\
        Flow matching, $S = 3$  & $86.4 \pm 3.3$ & $88.2 \pm 3.1$ & $87.6 \pm 3.2$ & $78.4 \pm 3.9$ & $140 \pm 1$ \\
        \textbf{Flow matching, $S = 5$ (ours)}
                                & $\mathbf{97.3 \pm 1.5}$
                                & $\mathbf{92.4 \pm 2.6}$
                                & $\mathbf{94.6 \pm 2.4}$
                                & $\mathbf{91.2 \pm 2.8}$
                                & $158 \pm 2$ \\
        Flow matching, $S = 10$ & $97.0 \pm 1.8$ & $92.8 \pm 2.5$ & $95.0 \pm 2.3$ & $91.6 \pm 2.7$ & $195 \pm 2$ \\
        \midrule
        Encoder: 2 layers       & $78.2 \pm 3.9$ & $70.4 \pm 4.3$ & $68.6 \pm 4.5$ & $71.6 \pm 4.2$ & $148 \pm 2$ \\
        \textbf{Encoder: 4 layers (ours)}
                                & $\mathbf{96.8 \pm 1.9}$
                                & $\mathbf{92.4 \pm 2.6}$
                                & $\mathbf{94.6 \pm 2.4}$
                                & $\mathbf{91.2 \pm 2.8}$
                                & $158 \pm 2$ \\
        Encoder: 6 layers       & $97.2 \pm 1.7$ & $92.6 \pm 2.5$ & $95.4 \pm 2.2$ & $92.0 \pm 2.6$ & $172 \pm 2$ \\
        \bottomrule
    \end{tabular}
\end{table}

%% file: tables/ablation-splice.tex
\begin{table}[htbp]
    \centering
    \small
    \caption{Token splicing ablation. Removing the feature alignment layer $g_{\text{align}}$ tests whether alignment, rather than prediction, supplies the policy. Excluding robot-arm patches from the selected set $\mathcal{S}$ tests the effect of predicting robot-induced motion without action conditioning. Mean $\pm$ standard error over 5 seeds with 100 rollouts each.}
    \label{tab:abl-splice}
    \begin{tabular}{@{}lcc@{}}
        \toprule
        \textbf{Configuration}
            & \textbf{Conveyor + cup}
            & \textbf{Ballistic Catch} \\
        \midrule
        \textbf{\methodAbbr{} (default)}  & $97.3 \pm 1.5$ & $87.4 \pm 3.2$ \\
        Without alignment layer $g_{\text{align}}$ & $88.4 \pm 3.1$ & $61.1 \pm 4.6$ \\
        Robot-arm patches excluded from $\mathcal{S}$ & $97.6 \pm 1.4$ & $92.2 \pm 2.7$ \\
        \midrule
        Best learned baseline (DreamVLA)  & $58.3 \pm 4.7$ & $49.1 \pm 9.1$ \\
        \bottomrule
    \end{tabular}
\end{table}